\documentclass{article}

\usepackage{times}
\usepackage{subfigure} 

\usepackage{natbib}

\usepackage{algorithm}
\usepackage{algorithmic}

\usepackage{hyperref}

\usepackage[accepted]{icml2016_arxiv}

\usepackage{url}
\usepackage[cmex10]{amsmath}
\usepackage{amssymb}  
\usepackage{graphicx,wrapfig}
\usepackage{bbm}
\usepackage{fixltx2e} 
\usepackage{array}
\usepackage{longtable}
\usepackage{tikz}
\usetikzlibrary{bayesnet}

\newcommand{\tightparagraph}[1]{\textbf{#1:}\ }

\newcommand{\indicator}[1]{\mathbbm{1}\{#1\}}

\newcommand{\varx}{\mathbf{x}}
\newcommand{\vary}{\mathbf{y}}

\newcommand{\varc}{\mathbf{z}}
\newcommand{\vars}{\mathbf{s}}

\newcommand{\varD}{D}
\newcommand{\funcf}{f}

\newcommand{\funcphi}{\phi}
\newcommand{\funcpsi}{\psi}
\newcommand{\funcchi}{\beta}
\newcommand{\funcind}{\mathbbm{1}}

\newcommand{\symloss}{\mathcal{L}}

\begin{document}

\twocolumn[
\icmltitle{Patterns for Learning with Side Information}


\icmlauthor{Rico Jonschkowski\footnote{}, Sebastian H\"ofer\footnotemark[1], Oliver Brock\\}
{\{rico.jonschkowski, sebastian.hoefer, oliver.brock\}@tu-berlin.de}
\icmladdress{Robotics and Biology Laboratory, Technische Universit\"at Berlin, Berlin, Germany}


\vskip 0.1in
]

%
%

\begin{abstract}

Supervised, semi-supervised, and unsupervised learning estimate a function given input/output samples. 
Generalization of the learned function to unseen data can be improved by incorporating \emph{side information} into learning. Side information are data that are neither from the input space nor from the output space of the function, but include useful information for learning it. 
In this paper we show that learning with side information subsumes a variety of related approaches, e.g.~multi-task learning, multi-view learning and learning using privileged information. Our main contributions are 
(i) a new perspective that connects these previously isolated approaches,
(ii) insights about how these methods incorporate different types of prior knowledge, and hence implement different \emph{patterns}, 
(iii) facilitating the application of these methods in novel tasks, as well as (iv) a systematic experimental evaluation of these patterns in two supervised learning tasks.
 

\end{abstract}


\vspace{-0.75cm}

\section{Introduction}
\label{sec: intro}

\footnotetext{The first two authors contributed equally to this work.}

An important branch of machine learning research focuses on supervised learning, estimating functions based on input/output samples with the goal of predicting the correct output for new inputs. However, generalization to unseen samples always requires prior knowledge (which we refer to as \emph{priors}) about the target function~\citep{mitchell_need_1980, schaffer_conservation_1994, wolpert_lack_1996}. By incorporating stronger priors, we can learn from less input/output samples or solve more challenging problems. But discovering useful priors that generalize over a wide range of tasks is difficult, especially if we only consider to define such priors over the target function, its input, and its output.

\begin{wrapfigure}[10]{r}{2.5cm}
\centering
\includegraphics[scale=0.5]{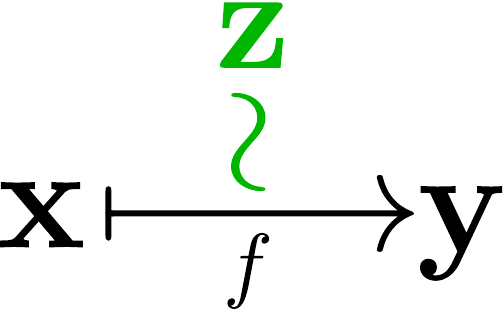}
\caption{Side information $\varc$ is related to function $\funcf(\varx)=\vary$.}\label{fig:context}
\end{wrapfigure}

For many problems, there are additional data $\varc$ available that are neither the input $\varx$ nor the output $\vary$ of function $\funcf$ but that carry valuable information about how $\funcf$ maps $\varx$ to $\vary$, as illustrated in Fig.~\ref{fig:context}.

We refer to this kind of data as \emph{side information}~\cite{chen_boosting_2012}, also known as privileged information~\cite{vapnik_new_2009}. Examples for side information are (i) intermediate results computed by the true underlying $\funcf$, (ii) output of a related function (with input $\varx$) that shares computations with $\funcf$, (iii) input of a related function (with output $\vary$) that shares computations with $\funcf$, or (iv) relations between inputs $\varx_i$ and $\varx_j$ or between outputs $\vary_i$ and $\vary_j$. 

\tightparagraph{Example} Suppose we want to estimate a function from the input/output samples: $3\!\mapsto\!14$, $\;5\!\mapsto\!30$, and $2\!\mapsto\!9$. From looking at these data, it is not immediately obvious what the true underlying function is.  However, if we provide side information and the prior that they correspond to intermediate values that $\funcf$ computes, in this case $3\!\mapsto\!\mathbf{9}\!\mapsto\!14$, $\;5\!\mapsto\!\mathbf{25}\!\mapsto\!30$, and $2\!\mapsto\!\mathbf{4}\!\mapsto\!9$, we see that the function first squares its input and then adds five to the intermediate result, $\funcf(\varx)=\varx^2 + 5$. Side information together with a prior about how they relate to $\funcf$ reveal the underlying function.

Incorporating priors about how $\varc$ relates to $\funcf$ is what we call \emph{learning with side information}. By enforcing consistency with these priors, we regularize learning which improves generalization. Note that we use side information \emph{only during training, not for prediction}. There are a number of approaches in the literature that (often implicitly) follow the paradigm of learning with side information and demonstrate impressive results. This paper connects these lines of work, makes the underlying paradigm explicit, and attributes the improved generalization to the use of priors enabled by side information.

\subsection{Prior Knowledge in Machine Learning}
\label{sec: background}

As mentioned before, machine learning incorporates priors about the target function $\funcf$ to generalize beyond observed data.
Although not always stated explicitly, priors about $\funcf$ are reflected in every component of a machine learning approach: in the \emph{hypothesis space} (e.g. by defining features, kernels, neural network structure), in the \emph{generation of training data} (e.g. by data augmentation), in the \emph{learning procedure} (e.g. by following a curriculum or decaying the learning rate), and in the \emph{learning objective} (e.g. by including a regularization loss).

Learning with side information provides an effective way to incorporate priors into the \emph{learning objective} by exploiting data that are neither input nor output data of the target function $f$ and are only required during training time. Note the difference to unsupervised and semi-supervised learning which only consider additional \emph{input} data.

\subsection{Contribution}
\label{sec: contribution}

This paper, for the first time, systematically analyzes how to exploit side information for improving generalization. It makes four main contributions:

The first contribution is a new perspective on machine learning problems. This perspective connects approaches from the literature such as~\emph{multi-task learning}, \emph{multi-view learning}, \emph{slow feature analysis}, \emph{learning using privileged information}, as well as several recent works in deep learning. By connecting these lines of work, which previously did not reference each other, we enable a new exchange of ideas between them. To facilitate communication, we provide a unifying formalization of learning with side information (Sec.~\ref{sec: cml}).

Our second contribution is a number of insights about these methods. First, they form a small set of \emph{patterns}~(Sec.~\ref{sec:patterns}) that correspond to different relationships between side information and target function (i-iv, second paragraph of the introduction). Second, the pattern's effectiveness in generalization is a result of incorporating priors about these relationships. Since patterns incorporate different priors, their effectiveness must depend on whether the learning task and side information match the prior. We, therefore, hypothesize that different patterns work for different tasks.

As our third contribution, we demonstrate how our insights advance learning with side information. 
First, we use the presented patterns to systematically compare different ways to use side information. 
Second, we present a new pattern that has not been studied in the literature (Sec. \ref{sec:multi-view_pattern}).
Third, we facilitate the practical application of learning with side information by giving a broad overview of successful applications in the literature (Appendix\footnote{The appendix can be found in the supplementary material.} \ref{app: table}) and by making our implementation publicly available\footnote{Our code for learning with side information is available at 
\url{https://github.com/tu-rbo/concarne}}.

The fourth contribution is a systematic experimental evaluation methods for learning with side information (Sec.~\ref{sec:experiments}). Our experiments confirm results from the literature by showing that learning with side information greatly improves generalization. 
Moreover, the results support our hypothesis from contribution two, showing that a pattern's performance strongly depends on the given task and the available side information.

\section{Learning with Side Information}
\label{sec: cml}

In \emph{learning with side information}, we estimate a function $\funcf: \varx \rightarrow \vary$ and optionally an auxiliary function $\funcchi$ 
by minimizing two objective functions, the \emph{main objective} $\symloss_\funcf$ and the \emph{side objective} $\symloss_\varc$:
\begin{align*}
	&\textrm{argmin}_{\funcf}\ \symloss_\funcf( \funcf\ |\ \{\varx_i,\vary_i\}_{i=1}^N), \\
	&\textrm{argmin}_{\funcf, \funcchi}\ \symloss_\varc( \funcf, \funcchi\ |\ \{\varx_i,\vary_i\}_{i=1}^N, \{\varc_j\}_{j=1}^{M}).
\end{align*}

To define $\symloss_\funcf$, we assume a supervised learning setting, in which the goal is to estimate a function $\funcf: \varx \rightarrow \vary$ from a set of $N$ input/output pairs $\{\varx_i,\vary_i\}_{i=1}^N$. Then, $\symloss_\funcf$ corresponds to a standard supervised learning objective, e.g.~mean-squared error for regression, and hinge loss for classification.

The side objective is captured by $\symloss_\varc$, which depends on \emph{side information} $\varc$ and can include the auxiliary function $\funcchi$. 
The exact form of $\symloss_\varc$, $\varc$ and $\funcchi$ depends on the \emph{pattern} applied (Sec.~\ref{sec:patterns}).
For all patterns, $\varc$ are data that are neither from the input space nor from the output space of $\funcf$ but carry valuable information about $\funcf$, and are only needed for learning, not for prediction.
Hence, the training data include $M$ side information samples in addition to the $N$ input/output pairs, $\varD$~$=$ $(\{\varx_i, \vary_i\}_{i=1}^N, \{\varc_j\}_{j=1}^{M})$. Each of the side information samples relates to one or more input/output samples, commonly $M=N$ 
or $M=N^2$.

To exploit $\varc$ for learning $\funcf$, we formulate priors about how $\varc$ relates to $\funcf$ in the side objective $\symloss_\varc$. To express $\symloss_\varc$, many patterns require 
$\funcf$ to be split into two functions, $\funcphi$ and $\funcpsi$, where $\funcphi$ maps $\varx$ to an intermediate representation $\vars$, and $\funcpsi$ predicts $\vary$ based on $\vars$, hence $\vary = \funcf(\varx) = \funcpsi(\funcphi(\varx)) = \funcpsi(\vars)$. This split exposes the representation $\vars$ and facilitates the formulation of $\symloss_\varc$ by relating $\vars$ and $\varc$, possibly using $\funcchi$. 
Often it allows us to omit $\funcpsi$ and $\vary$ from  $\symloss_\varc$, i.e. to define $\symloss_\varc( \funcphi, \funcchi\ |\ \{\varx\}, \{\varc\})$.
For example, in the multi-task pattern
(Sec.~\ref{sec:multi-task_pattern}) the intermediate representation $\vars$ is shared amongst the main task of predicting $\vary$ with function $\funcpsi(\vars)$ and an auxiliary task of predicting $\varc$ with $\funcchi(\vars)$. The auxiliary task regularizes the shared function $\funcphi$ and improves generalization for the main task.

Note that we intentionally kept this formalization narrow to improve readability. It is straightforward to extend the ideas presented here to a reinforcement learning setting, to multiple types of side information, to multiple intermediate representations, and to more than one side objective. 

\subsection{Training Procedures}
\label{subsec: training procedures}

Since learning with side information requires us to optimize multiple learning objectives affected by different subsets of training data and functions, we need appropriate training procedures. We have identified three common training procedures that differ with respect to the order in which they (i) optimize the two objectives and (ii) modify the functions $\funcf$ and $\funcchi$:

\textbf{Simultaneous learning} jointly trains $\funcf$ and $\funcchi$ by optimizing a weighted sum of the two learning objectives $\symloss_\funcf$ and $\symloss_\varc$~\citep{weston_deep_2012}. 
This procedure introduces the need to find a good weighting of the different learning objectives, which might be difficult if the gradients of the objectives differ by orders of magnitude and vary during learning.

If we split $\funcf$ into $\funcphi$ and $\funcpsi$, as described in the previous section, we can choose among two additional procedures. 
In the \textbf{decoupled procedure}, we first optimize the side objective $\symloss_\varc( \funcphi, \funcchi\ |\ \{\varx\}, \{\varc\})$, while adapting $\funcphi$ and $\funcchi$ to learn the intermediate representation $\vars$.
Then, we optimize the main objective $\symloss_\funcf( \funcphi, \funcpsi\ |\ \{\varx,\vary\})$, while keeping $\funcphi$ (and $\funcchi$) fixed. This simple procedure is only applicable if the side objective provides enough guidance to learn a task-relevant representation $\vars$, whereas the simultaneous procedure is also applicable for ``weak'' side objectives $\symloss_\varc$. 
To alleviate this problem, the \textbf{pre-train and finetune procedure} first applies the decoupled procedure, but then optimizes $\symloss_\funcf( \funcphi, \funcpsi\ |\ \{\varx,\vary\})$ while adapting $\funcphi$, too, in order to fine-tune $\vars$ for the task.
This strategy is popular in deep learning as unsupervised pre-training \citep{erhan_why_2010} and can be applied analogously for learning with side information. For this procedure to have an effect, $\symloss_\funcf$ must not be convex (otherwise, the pre-training step would be unlearned). 


\section{Patterns for Learning with Side Information}
\label{sec:patterns}

We will now present different approaches for learning with side information, which we have grouped into patterns. We describe for each pattern the general idea, the underlying prior, the side information $\varc$, the side objective $\symloss_\varc$, and the auxiliary function $\funcchi$. We point to successful applications of each pattern (summarized in Appendix~\ref{app: table}) and visualize the patterns with schemas as in Fig.~\ref{fig:pattern_direct}. 


\tightparagraph{How to read the schemas} The schemas represent computation flow graphs where functions (drawn as arrows) connect variables (represented as nodes), both of which follow the definitions from Section~\ref{sec: cml}. Predictions of variables are indicated by $\hat{\cdot}$. The target function is depicted in black. Additional elements that are only required at training time and can be omitted during prediction are shown in gray, except for side information and the corresponding learning objectives, which are highlighted in green. Learning objectives are visualized by connecting variables with $\sim$ to denote that the objective enforces similarity between these variables. The $=$-sign (see Fig.~\ref{fig:pattern_pairwise}) indicates that a function is replicated (e.g. by weight sharing).

Note that these graphs are \emph{not} probabilistic graphical models (PGMs). We provide PGMs as a complementary visualization of causal dependencies in Appendix~\ref{app: pgm}. In contrast, the computation flow graphs are advantageous for the purpose of this paper since (i) they discriminate between variables and functions, (ii) they expose the sequence of computation, (iii) they visualize the learning objectives, and thus (iv) are easily converted into neural networks, which are employed by most of the related works reviewed in this paper.



\subsection{Direct Pattern}
\label{sec:direct_pattern}

\begin{wrapfigure}[5]{r}{3.5cm}
\vspace{-0.5cm}
\centering
\includegraphics[scale=0.3]{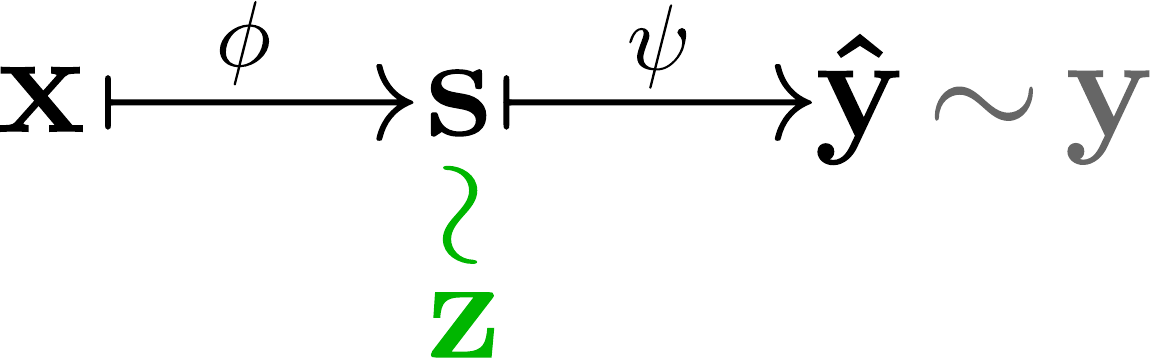}
\caption{Direct pattern}\label{fig:pattern_direct}
\end{wrapfigure}

The direct pattern leverages known, intermediate results of the computation performed by $\funcf$. Given these intermediate results as side information $\varc$, we can learn a function $\funcphi$ that transforms $\varx$ into the representation $\vars$ such that $\vars \sim \varc$, as shown in Fig.~\ref{fig:pattern_direct}. No auxiliary function $\funcchi$ is required. The pattern is only applicable if $\varc$ makes it easier to predict $\vary$, and if $\varx$ contains enough information to predict $\varc$. The example in Section~\ref{sec: intro} is an instance of this pattern.

To formalize this pattern, we use a suitable supervised learning side objective $\symloss_\varc = \mathcal{L}_\textrm{direct}(\funcphi\ |\ \{ \varx, \varc \} )$ that enforces the representation $\vars$ to be equal to the side information $\varc$.


\tightparagraph{Applications} Machine learning approaches in computational biology frequently use this pattern to combine understanding from biology research with learning. For example, in contact prediction, the goal is to predict which parts of a folded protein are close to each other based on the DNA sequence that describes the protein. Virtually all learning-based approaches to this problem first predict intermediate representations $\vars$, such as secondary structures (local 3D structure categories), and then use $\vars$ to predict contacts~\citep{cheng_improved_2007}. The representation $\vars$ can be reliably estimated which greatly facilitates learning $\funcphi$.


\emph{Knowledge transfer} \citep{vapnik_learning_2015} uses this pattern, but includes an additional step of extracting features $\funcchi(\varc)$ from the side information. Function $\funcphi$ is then learned by regression, such that $\vars \sim \funcchi(\varc)$. They also suggest augmenting $\vars$ with the original input $\varx$. Similarly, \citet{chen_boosting_2012} suggest to reconstruct only highly predictive features of $\varc$ using a modified version of AdaBoost.

\subsection{Multi-Task Pattern}
\label{sec:multi-task_pattern}

\begin{wrapfigure}[5]{r}{3.5cm}
\vspace{-0.6cm}
\centering
\includegraphics[scale=0.3]{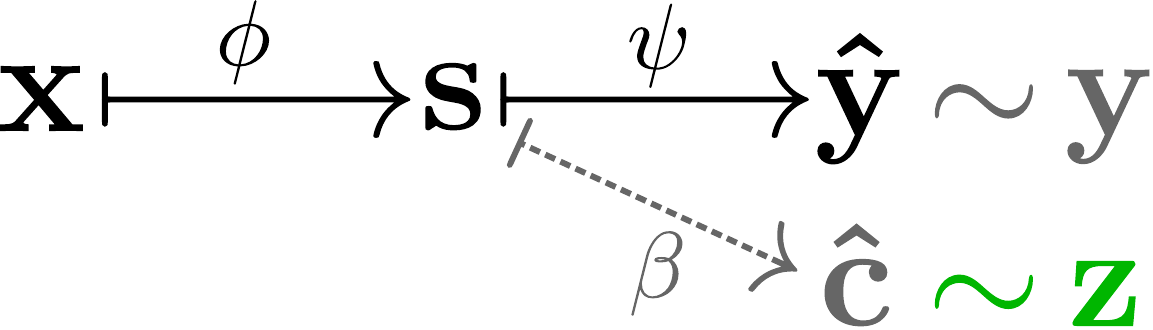}
\caption{Multi-task pattern}\label{fig:pattern_multitask}
\end{wrapfigure}

This pattern applies when the side information $\varc$ are outputs of a related function (with input $\varx$) that shares computations with the function we want to estimate. As illustrated in Fig.~\ref{fig:pattern_multitask}, the pattern assumes that the target function $\funcf =  \funcpsi \circ \funcphi$ and the related function $\funcchi \circ \funcphi$ share $\funcphi$ and therefore have the same intermediate representation $\vars = \funcphi(\varx)$. By training the representation to predict both $\vary$ using $\funcpsi$, and $\varc$ using the auxiliary learnable function $\funcchi: \vars \rightarrow \varc$, we incorporate the prior that \emph{related tasks share intermediate representations.} This pattern corresponds to multi-task learning \citep{caruana_multitask_1997}, a type of transfer learning~\citep{pan_survey_2010}.

To apply the multi-task pattern, we can use any suitable learning objective from supervised learning in order to learn to predict $\varc$ from $\varx$, i.e.~ $\symloss_\varc=\mathcal{L}_\textrm{multi-task} (\funcphi, \funcchi \ |\ \{ \varx, \varc \})$. 

\tightparagraph{Applications} Multi-task learning has been successfully applied in a wide variety of tasks~\citep{caruana_multitask_1997,pan_survey_2010}. Recently, \citet{zhao_improved_2015}~proposed to use object pose information to improve object recognition in a convolutional deep neural network. 
Similarly, \citet{levine_end-end_2015}~use image classification and pose prediction as side information to teach a robot remarkable vision-based manipulation skills, such as stacking lego blocks or screwing caps onto bottles. 


\subsubsection{Irrelevance Pattern}
\label{subsec: irrelevance}

\begin{wrapfigure}[6]{r}{4.5cm}
\vspace{-0.6cm}
\centering
\includegraphics[scale=0.3]{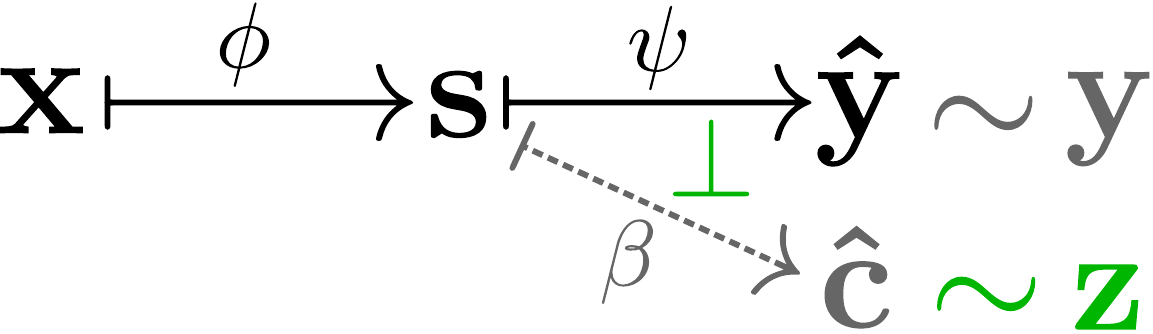}
\caption{Exploiting irrelevant side information.}
\label{fig:irrelevant}
\end{wrapfigure}

A special case of the multi-task patterns exploits knowledge about unrelated tasks, by enforcing the prediction of the side information to be orthogonal to the main task~\cite{romera-paredes_exploiting_2012}. This idea is formalized by forcing $\funcpsi$ to be orthogonal to the auxiliary prediction function $\funcchi$ (see Fig.~\ref{fig:irrelevant}), which allows to use knowledge about irrelevant distractors present in the input data. However, it is unclear how to efficiently formulate the orthogonality constraints between $\funcpsi$ and $\funcchi$ for the non-linear case.




\subsection{Multi-View Pattern}
\label{sec:multi-view_pattern}

The multi-view pattern is complementary to the multi-task pattern, treating side information as input instead of output. It applies when $\varc$ are inputs of a related function (with output~$\vary$) that share computations with $\funcf$. This pattern corresponds to multi-view learning~\citep{sun_survey_2013}. 

\begin{wrapfigure}[12]{r}{3.5cm}
\vspace{-0.8cm}
\centering
\subfigure[Labeled $\varx$ data]{\includegraphics[scale=0.3]{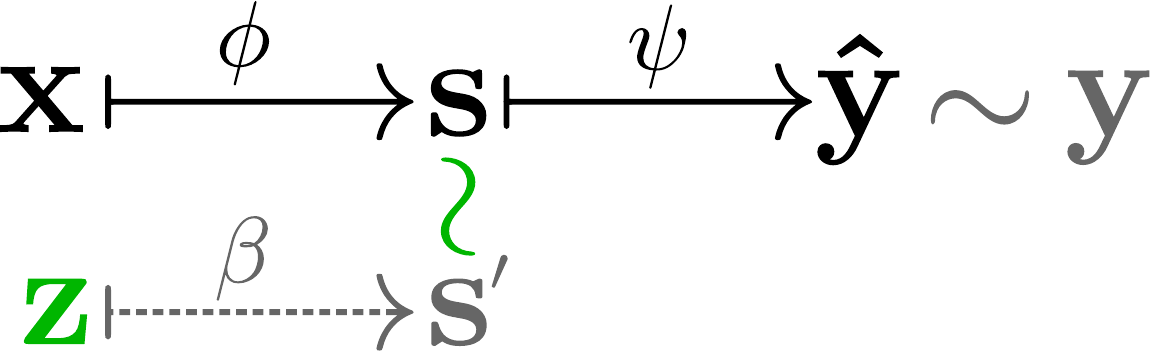}}
\\
\subfigure[Labeled $\varc$ data]{\includegraphics[scale=0.3]{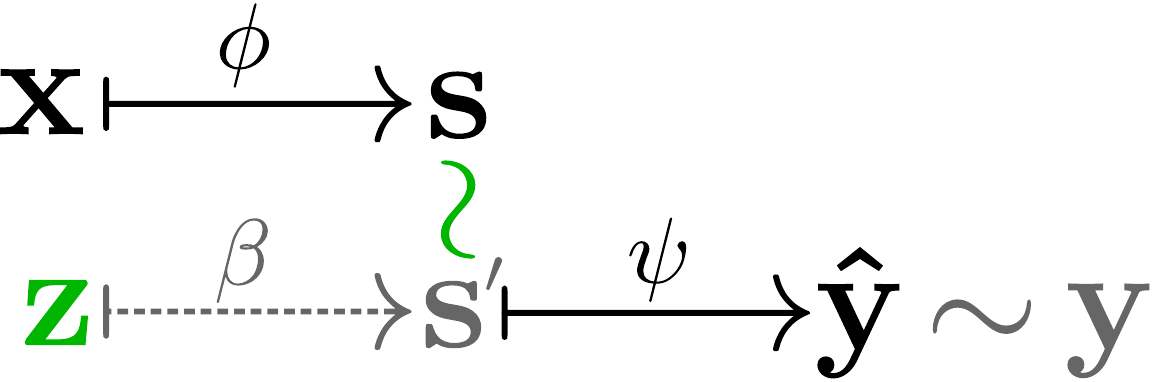}}
\caption{Multi-view (correlation) pattern}\label{fig:pattern_correlation}
\end{wrapfigure}

\begin{wrapfigure}[7]{r}{3.5cm}
\vspace{-0.2cm}
\centering
\includegraphics[scale=0.3]{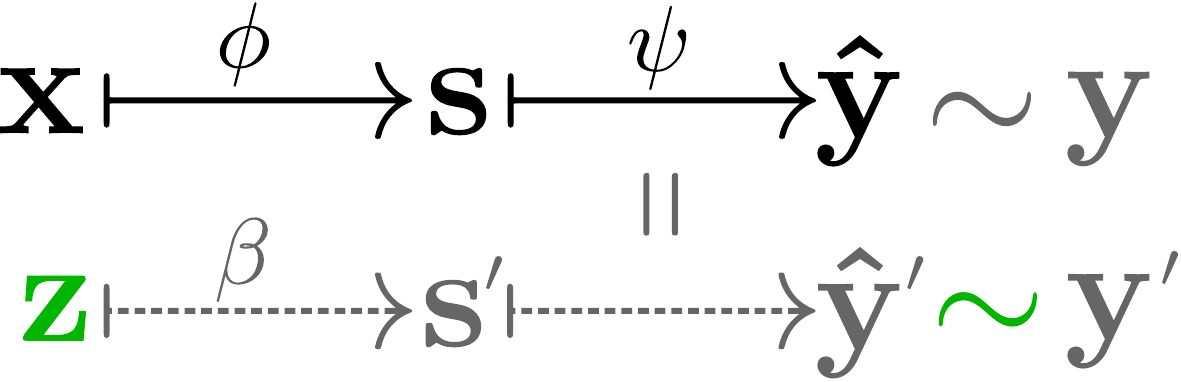}
\caption{Multi-view prediction pattern}\label{fig:pattern_shared_prediction}
\end{wrapfigure}

When we treat $\varc$ as auxiliary input, we can use it in two different ways: explicitly by \emph{correlating} it with the original input $\varx$ (Fig.~\ref{fig:pattern_correlation}), or implicitly by \emph{predicting} the target output (Fig.~\ref{fig:pattern_shared_prediction}).
In both cases, we learn functions $\funcphi: \varx \mapsto\! \vars$ and $\funcchi: \varc \mapsto\! \vars'$, such that~$\vars \sim \vars'$.



The \textbf{multi-view (correlation) pattern} assumes that \emph{correlated representations computed from related inputs are a useful intermediate representation for predicting the target output.}
It can be formalized with a learning objective that enforces the correlation between $\funcphi(\varx)$ and $\funcchi(\varc)$, e.g. the mean squared error 
$\symloss_\varc=\mathcal{L}_\textrm{multi-view}(\funcphi, \funcchi\ |\ \{ \varx, \varc \} ) = \sum_i ||\funcphi(\varx_i) - \funcchi(\varc_i)||^2$. If we apply the decoupled training procedure, i.e.~only optimize the  objective, we have to add constraints, e.g. unit variance, to $\mathcal{L}_\textrm{multi-view}$ in order to avoid the trivial solution of having a constant intermediate representation. In case $\funcphi$ and $\funcchi$ are linear, $\mathcal{L}_\textrm{multi-view}$ with unit variance corresponds to Canonical Correlation Analysis (CCA).

\tightparagraph{Applications} The pattern is often employed in multi-modal scenarios~\citep{sun_survey_2013}. 
\citet{chen_recognizing_2014} show how to enhance object recognition from RGB-only images by leveraging depth data as side information during training. In computational neuroscience, the pattern is widely used to learn from multiple modalities (e.g., EEG and fMRI) or across subjects~\citep{dahne_finding_2014}. 
The pattern can also be applied for clustering~\citep{feyereisl_privileged_2012}. The idea is to repeatedly cluster on both $\{\varx_i\}$ and $\{\varc_j\}$ and then return the clustering of $\varx$ with the highest agreement with $\varc$.
In a recent article, \citet{wang_deep_2015} suggest and compare deep architectures that combine multi-task and multi-view learning, and show that a deep canonically correlated auto-encoder gives superior results for visual, speech, and language learning.




The \textbf{multi-view prediction pattern} is based on the prior that \emph{predicting the target output from related inputs requires similar intermediate representations.} It trains the functions $\funcphi: \varx \mapsto\! \vars$ and $\funcchi: \varc \mapsto\! \vars'$ such that both $\vars$ and $\vars'$ map to the target output using the same prediction function $\funcpsi$, e.g. using weight sharing. Since $\vars$ and $\vars'$ are coupled to $\vary$ via the main objective, we do not only regularize $\funcphi$, but also $\funcpsi$.

Despite their similarities, we are not aware of any systematic comparison of multi-view and multi-task learning. Neither have we found applications of the prediction pattern in the literature. Our experiments provide a first empirical comparison of these patterns~(Sec.~\ref{sec:experiments}). 


\subsection{Pairwise Patterns}
\label{sec:pairwise_pattern}

\begin{wrapfigure}[6]{r}{3.7cm}
\vspace{-0.6cm}
\centering
\includegraphics[scale=0.3]{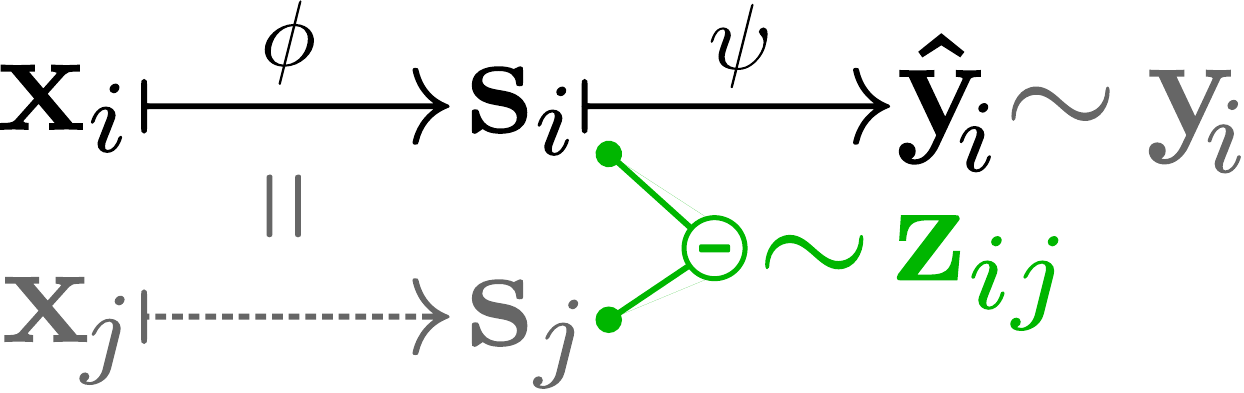}
\caption{Pairwise pattern}\label{fig:pattern_pairwise}
\end{wrapfigure}

Pairwise patterns use side information $\varc_{ij}$ that carry information about the relationship between samples $i$ and $j$ to shape the intermediate representation, e.g. the difference between their intermediate representations (Fig.~\ref{fig:pattern_pairwise}, the $=$ indicates weight sharing). 

\subsubsection{Pairwise Similarity/Dissimilarity Pattern}
\label{sec:similarity}

If the side information gives information about similarity of samples with respect to the task, we can impose the prior that \emph{samples that are similar (dissimilar) according to their side information should have similar (dissimilar) intermediate representations.} Such side information is often available as information about local neighborhoods of samples~\citep{tenenbaum_global_2000}. Another powerful source of similarity information are time sequences, since temporally subsequent samples often have similar task-relevant properties, as exploited by slow feature analysis (SFA) and temporal coherence~\citep{wiskott_slow_2002,weston_deep_2012}. Additionally, an intelligent teacher can provide information about which samples are similar~\citep{vapnik_learning_2015}.


Similarity can be enforced with a squared loss on the distance between similar samples:
\begin{align*}
\mathcal{L}_\text{sim.}(\funcphi \mid \{ \varx, \varc \} )
 = \sum_{i,j} || \funcphi(\varx_i) - \funcphi(\varx_j) ||^2\ \funcind(\varc_{ij} = {\text{sim.}}),
\end{align*}
where $\funcind$ denotes the indicator function. Solely using this objective might lead to trivial solutions where all samples are mapped to a constant. We can resolve this problem by imposing additional balancing constraints on $\vars$~\citep{weston_deep_2012} or selectively push samples apart that are dissimilar according to the side information (or optionally according to the labels $\vary$):
\begin{align*}
\mathcal{L}_\text{dis.}(\funcphi \mid \{ \varx, \varc \} )
 = \sum_{i, j} \sigma(||\funcphi(\varx_i) - \funcphi(\varx_j)||)\ \funcind(\varc_{ij} = \text{dis.}),
\end{align*}
where $\sigma$ 
is a function that measures the proximity of dissimilar samples in representation $\vars$. Candidates for $\sigma$ are the margin-based $\sigma(d)=\max(0,m-d^2)$ for some pre-defined margin $m$ \citep{hadsell_dimensionality_2006}, the exponential of the negative distance $\sigma(d)=e^{-d}$ \citep{jonschkowski_state_2014}, or the Gaussian function $\sigma(d)=e^{-d^2}$ \citep{jonschkowski_learning_2015}.
Another way to avoid trivial solutions is to impose an input-reconstruction objective, e.g. by using an auto-encoder~\cite{watter_embed_2015}.

\citet{vapnik_learning_2015} incorporate similarity information into support vector machines by replacing the free slack variables with a function of $\varc$. This method incorporates the prior that slack variables should be similar for samples with similar side information.


\tightparagraph{Applications} 
This pattern has been shown to successfully guide the learner in identifying task-relevant properties of $\varx$. 
\citet{hadsell_dimensionality_2006}~show how to learn a lighting invariant pose representation of objects in the NORB dataset. \citet{weston_deep_2012}~show that regularizing a convolutional network with a temporal coherence objective outperforms pure supervised object classification in the COIL-100 dataset by $20\%$ in terms of recognition accuracy. 

Recent works show how to apply this pattern to reinforcement learning settings.
\citet{watter_embed_2015} exploit the time sequence to jointly learn a state representation and the world dynamics from raw observations for a variety of standard tasks, such as cart-pole balancing.
\citet{jonschkowski_learning_2015} apply the pattern in a robot navigation task, and show how leveraging temporal and robot action information enable the robot to learn a state representation from raw observations, despite the presence of visual distractors.


Note that this pattern only preserves \emph{local} similarities between samples. 
If the side information provides a global distance metric, \citet{weston_deep_2012} propose to formulate side objectives for learning a distance-preserving mapping of $\varx$ to $\varc$, e.g. based on multi-dimensional scaling~\citep{kruskal_multidimensional_1964}. Alternatively, the distance metric can be learned using side information~\citep{fouad_incorporating_2013}.


\subsubsection{Pairwise Transformation Pattern}


Instead of exploiting only binary similarity information between samples, the pairwise transformation pattern exploits continuous information about the relative transformations between samples, to make \emph{the internal representation (or parts of it) consistent or equivariant with the known relative transformations.}  Such side information is often available in robot and reinforcement learning settings.

\begin{wrapfigure}[25]{r}{4.3cm}
\vspace{-0.6cm}
\centering
\subfigure[Predicting transformed input\label{fig:transf_pred_input}]{\includegraphics[scale=0.3]{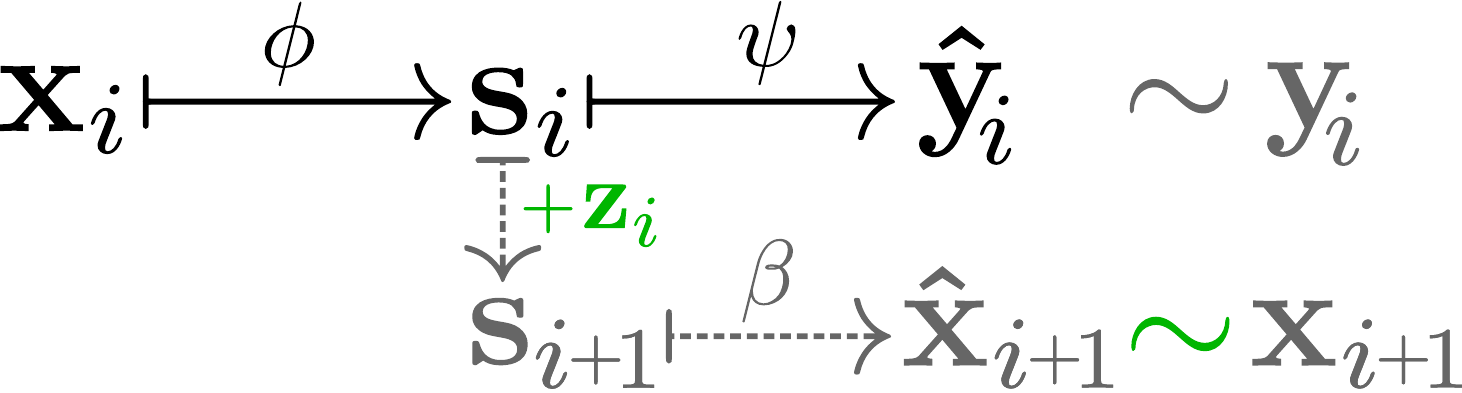}}
\\
\subfigure[Predicting representation\label{fig:transf_pred_state}]{\includegraphics[scale=0.3]{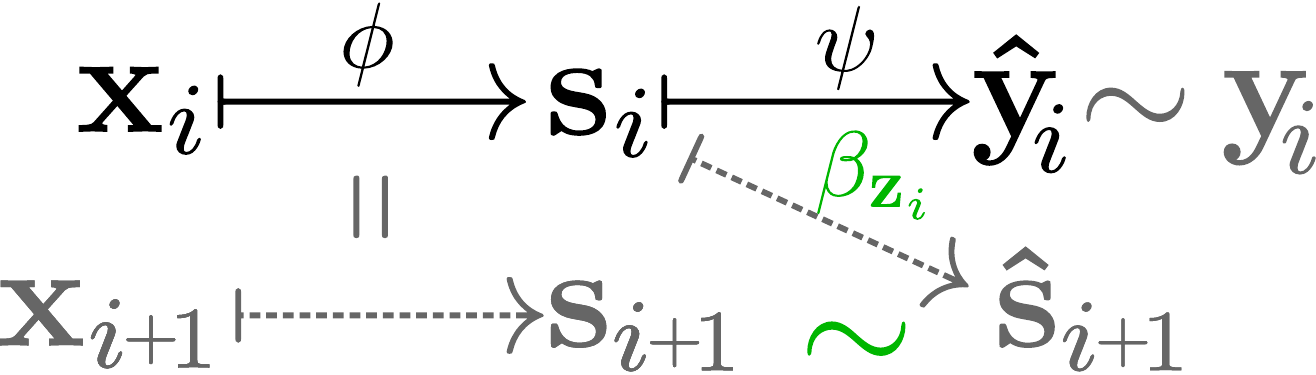}}
\\
\subfigure[Predicting transformation\label{fig:transf_pred_transf}]{\includegraphics[scale=0.3]{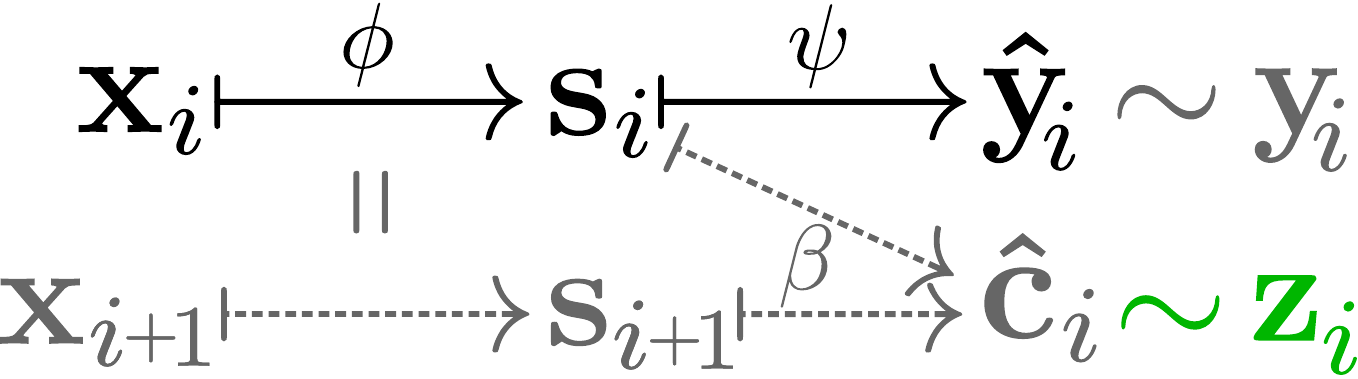}}
\\
\subfigure[Comparing pairs of transformations\label{fig:transf_pairs}]{
\includegraphics[scale=0.3]{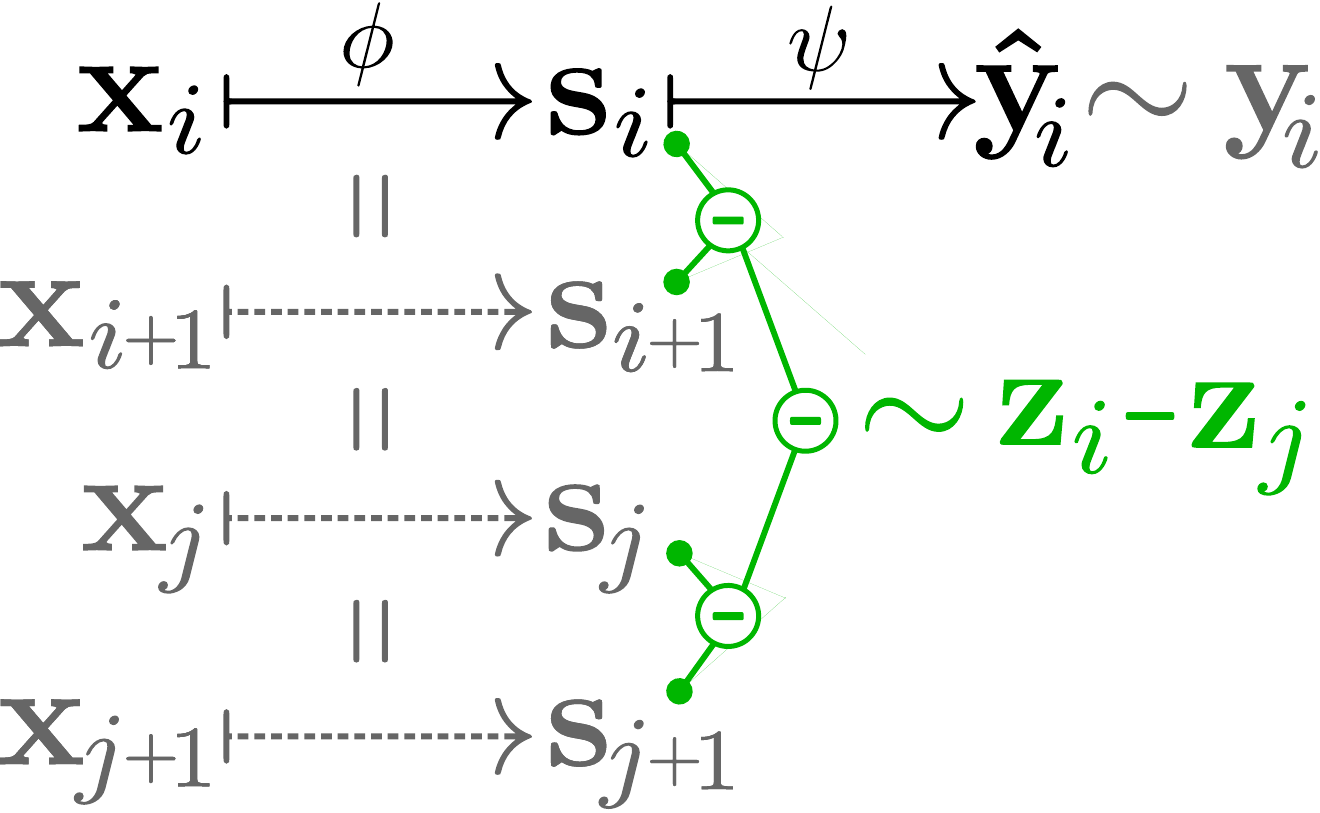}
}
\caption{Pairwise transf. patterns}
\end{wrapfigure}

Consistency with the transformations $\varc$ can be enforced in different ways: (a) \citet{hinton_transforming_2011} require the transformation $\varc$ to affect $\vars$ in a known way, and suggest the \emph{transforming autoencoder model} shown in Fig.~\ref{fig:transf_pred_input} to learn such an $\vars$. The idea is to learn to reconstruct the transformed input from the original input and the known transformation. 
(b) If the transformations in $\vars$ are unknown, \citet{jayaraman_learning_2015} suggest to learn these transformations as an auxiliary task using the pattern depicted in Fig.~\ref{fig:transf_pred_state}. (c) We can also turn this approach around and try to predict the transformation based on the original and the transformed representation~\citep{agrawal_learning_2015} as depicted in Fig.~\ref{fig:transf_pred_transf}. All three variants (a)-(c) enforce equivariance of $\vars$ with respect to the relative transformations, and can be trained using supervised side objectives.
(d) Instead of optimizing for equivariance, we can also enforce that the same transformation has the same effect, when applied to different samples (Fig.~\ref{fig:transf_pairs}). When transformations are discrete, we formalize this by penalizing the squared difference of the change in internal representation after applying the same transformation:
\begin{align*}
\label{eq:transformation}
\mathcal{L}_\text{transf.}(\funcphi \mid \{ \varx, \varc \} ) 
 = \sum_{\mkern+0mu i,j} \mkern-0mu ||\Delta\funcphi(\varx_i)& - \Delta\funcphi(\varx_j)||^2 \funcind(\varc_i \mkern-6mu = \varc_j), 
\end{align*}
where $\Delta$ denotes the change caused by the transformation, i.e. $\Delta\funcphi(\varx_i) = \funcphi(\varx_{i+1}) - \funcphi(\varx_i)$ for sequential data. This objective can be extended to continuous transformations by replacing the indicator function with a similarity function $\sigma(\varc_i \mkern-6mu - \varc_j)$ from Section~\ref{sec:similarity}. Variants of this pattern allow to enforce only locally consistent transformations, by multiplying $\sigma(\funcphi(\varx_i) - \funcphi(\varx_j))$, or to enforce only consistent magnitudes of change by comparing norms $||\Delta\funcphi(\varx_i)||$ \citep{jonschkowski_learning_2015}.

\tightparagraph{Applications} Many results in the literature demonstrate the usefulness of the pairwise transformation pattern. 
\citet{agrawal_learning_2015}~report that using relative pose information as side information can reduce the error rate on MNIST by half with respect to pure supervised learning. They also demonstrate the approach for scene recognition on the SUN dataset, and show that pre-training using limited of relative pose side information is almost as good class-based supervision.
\citet{jayaraman_learning_2015}~demonstrate a recognition accuracy of $\approx 50\%$ on the KITTI dataset, outperforming pure supervised learning ($41.81\%$ accuracy) and SFA ($47.04\%$). 
Interestingly, both works enforce learning a pose \emph{equivariant} representation, although the classification task they address requires \emph{invariance}. It is still unclear why equivariant representations help in such tasks~\citep{lenc_understanding_2014}.




\subsubsection{Label Distance Pattern}
\label{subsubsec: label distance}

\begin{wrapfigure}[4]{r}{4cm}
\vspace{-0.3cm}
\centering
\includegraphics[scale=0.3]{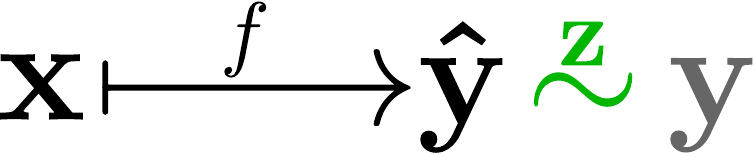}
\caption{Label distances}\label{fig:pattern_pairwise_label}
\end{wrapfigure}

The label distance pattern is a special case of the pairwise pattern, where the side information defines distances between labels, not samples (see Fig.~\ref{fig:pattern_pairwise_label}). 
An instance of this pattern, often used in structured prediction, is hierarchical multi-class learning~\citep{silla_jr_survey_2010}, where a hierarchy is imposed on the labels to penalize misclassifications between samples with similar classes less severely.

\section{Experiments}
\label{sec:experiments}

The related work, discussed in the previous section, demonstrated that learning with side information greatly improves generalization. In our experiments, we provide, for the first time, a systematic comparison of the different patterns in two supervised learning tasks. We outline the experimental rationale and results here, and refer to Appendix~\ref{app: experiments} for details.


\subsection{Synthetic Task}
\label{exp: synthetic task}

\begin{figure}[H]
\vspace{-0.1cm}
\center
\subfigure[Direct side information\label{fig: toy direct}]{\includegraphics[width= 0.8\columnwidth]{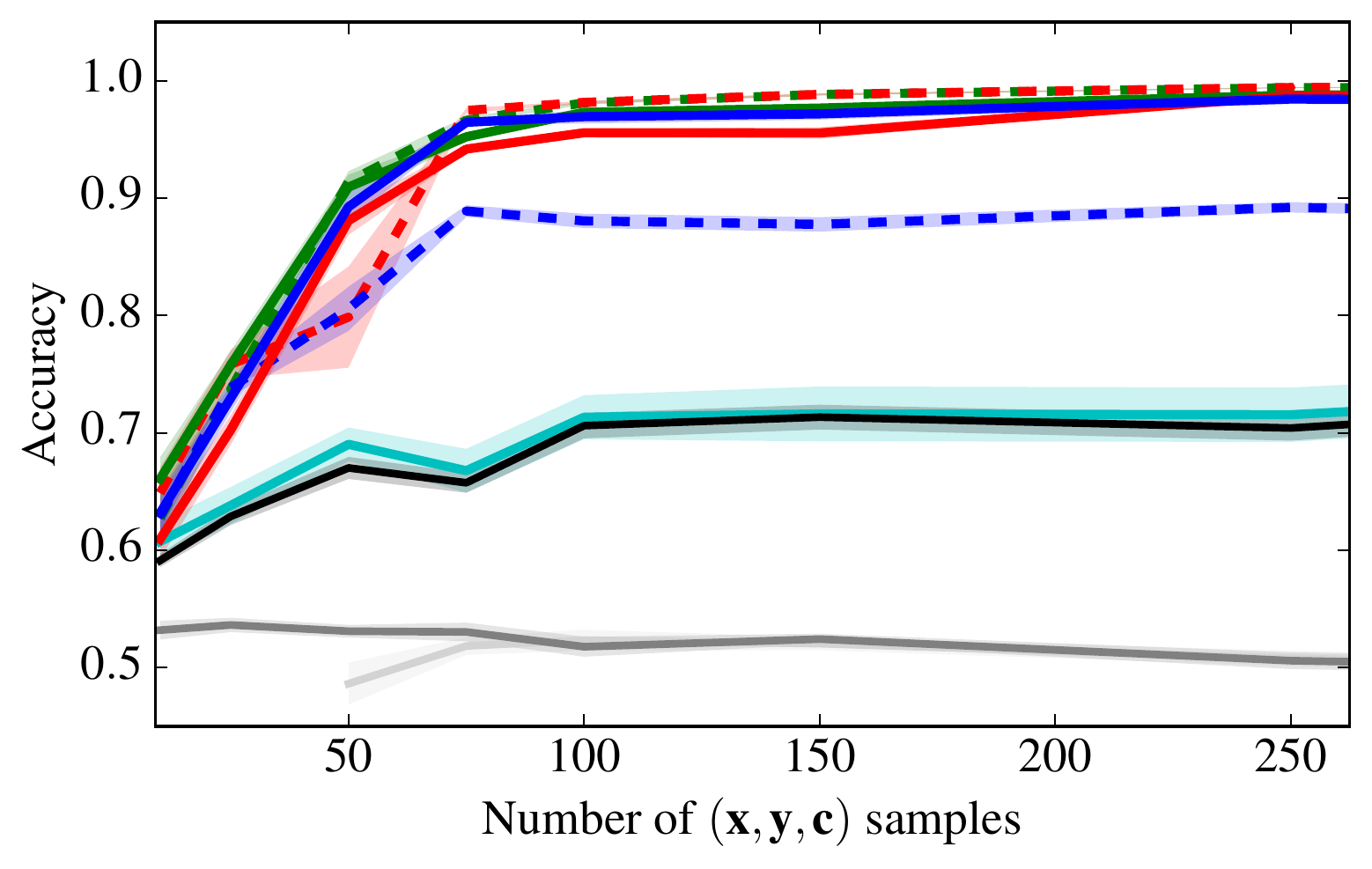}}
\\
\subfigure[Embedded side information \label{fig: toy embedded}]{\includegraphics[width= 0.8\columnwidth]{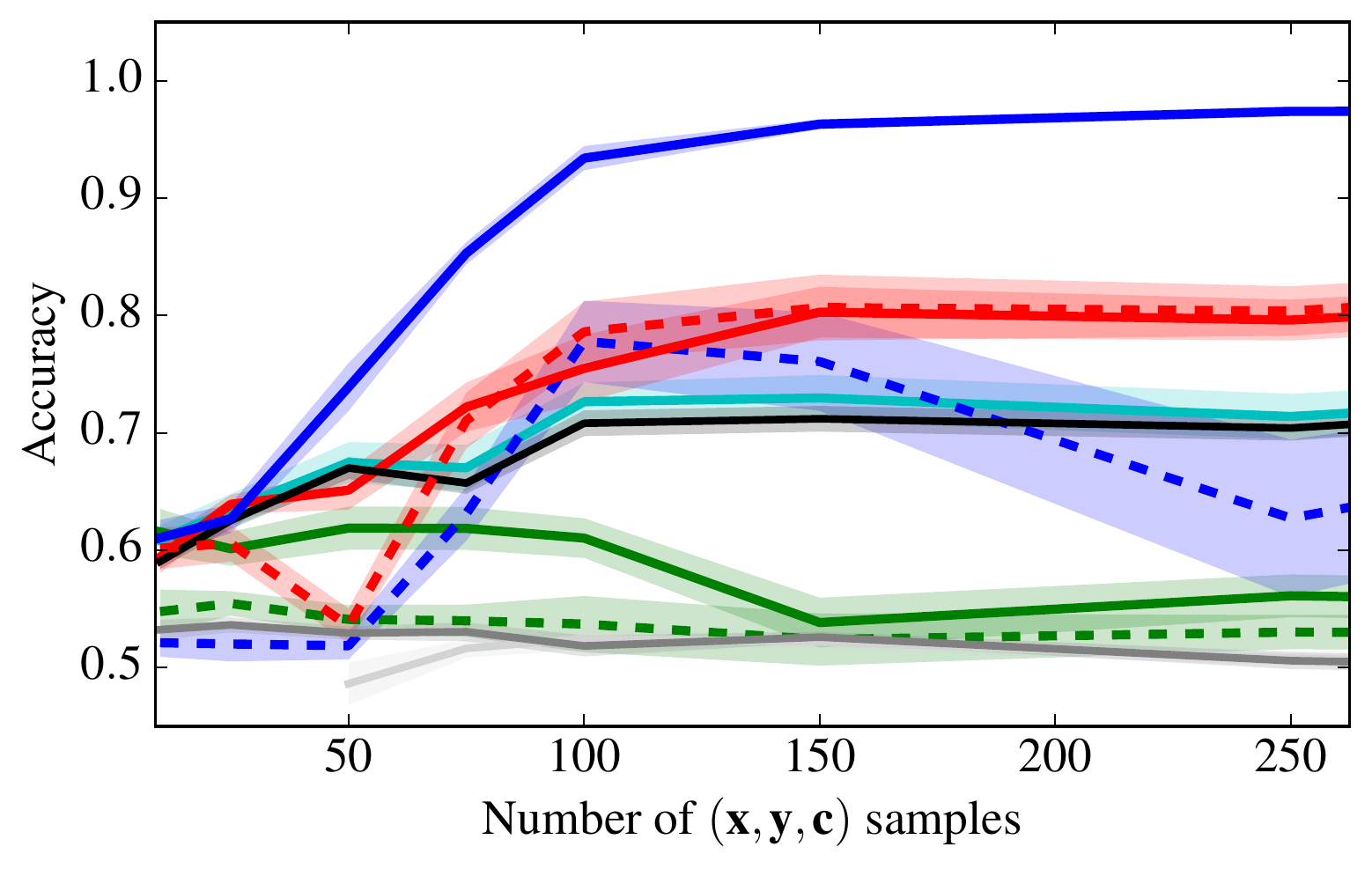}}
\\
\subfigure[Pairwise side information \label{fig: toy relative}]{\includegraphics[width= 0.8\columnwidth]{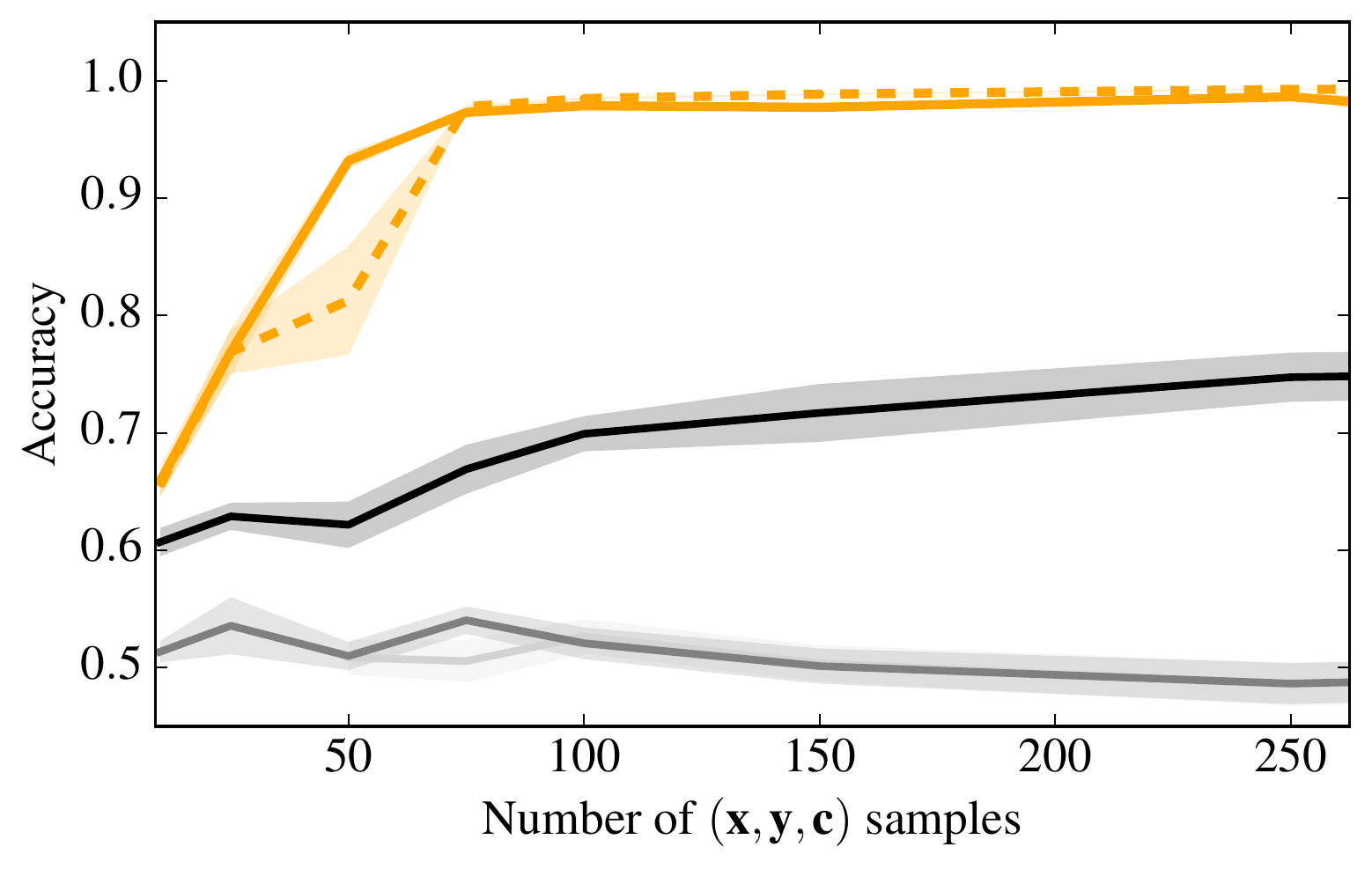}}
\\
\subfigure{\includegraphics[width=0.8\columnwidth]{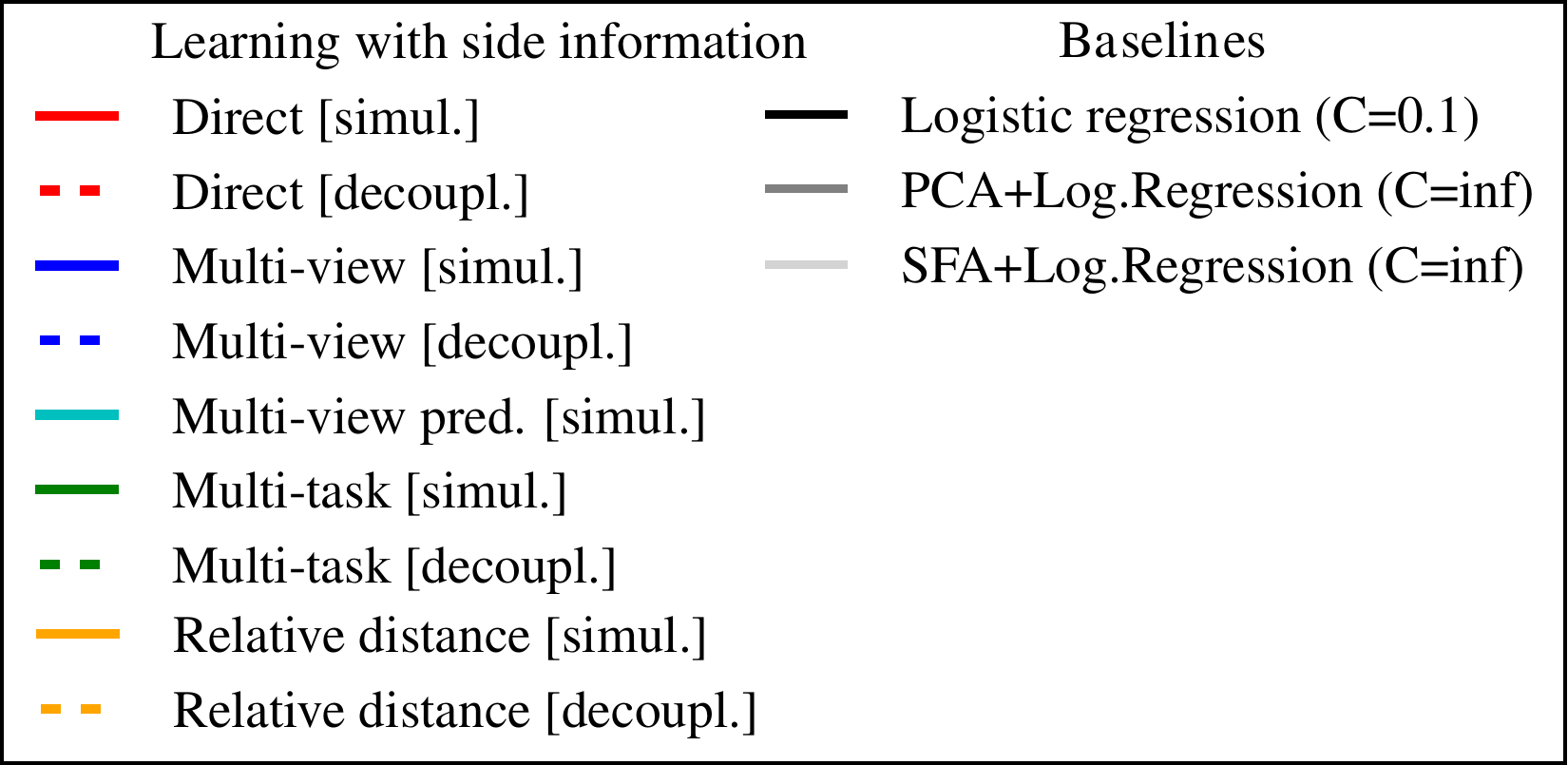}}
\caption{Results for the synthetic task, averaged over 10 runs. The most suitable patterns outperform supervised and semi-supervised baselines and generalize better with much less data.}
\label{fig:toy results}
\end{figure}

In the first, synthetic, experiment the goal is to predict the position of a randomly moving agent in a 1-dimensional state space $\vars$. The learner cannot perceive $\vars$ directly, but gets an observation $\varx$, which embeds $\vars$ and a set of distractor signals in a high-dimensional space. 
The learned functions are linear ($\funcphi$ and $\funcchi$), and logistic functions ($\funcpsi$), respectively.
We study the effect of different combinations of (i)~side information, (ii)~patterns, and (iii)~training procedures on prediction accuracy. The side information are either  a noisy variant of the real state (\emph{direct} side information, Fig.~\ref{fig: toy direct}), a second noisy high-dimensional observation (\emph{embedded}, Fig.~\ref{fig: toy embedded}), or a noisy variant of the agent's actions, i.e.~the agent's relative motion (\emph{pairwise}, Fig.~\ref{fig: toy relative}). We apply the direct, multi-view, multi-task, and pairwise transformation patterns, and perform training using the decoupled and simultaneous procedure (pre-training is futile with linear functions). We compare to supervised and semi-supervised baselines.


\tightparagraph{Results}
While none of the baselines are able to solve the task with the given amount of training data, for each form of side information, at least one pattern achieves close to optimal performance. For the simple direct side information (Fig.~\ref{fig: toy direct}) all patterns except the multi-view prediction pattern are applicable; the reason is that the direct data correspond to the real state, and thus make solving the task almost trivial.
This does not hold true for the embedded side information~(Fig.~\ref{fig: toy embedded}). Here, the simultaneously trained multi-view correlation pattern clearly outperforms all other methods. The reason is that the embedded side information exactly matches the prior of the multi-view pattern. The pairwise transformation pattern, when applied to pairwise side information (Fig.~\ref{fig: toy relative}), is as effective as learning from direct side information. 
Overall, the experiments confirm our hypothesis that the effectiveness of each pattern strongly depends on the type of side information.

\subsection{Handwritten Character Recognition}

\begin{figure}[H]
\subfigure{\includegraphics[width=0.085\columnwidth]{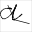}}
\hfill
\subfigure{\includegraphics[width=0.085\columnwidth]{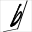}}
\hfill
\subfigure{\includegraphics[width=0.085\columnwidth]{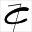}}
\hfil
\subfigure{\includegraphics[width=0.085\columnwidth]{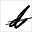}}
\hfill
\subfigure{\includegraphics[width=0.085\columnwidth]{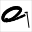}}
\hfill
\subfigure{\includegraphics[width=0.085\columnwidth]{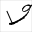}}
\hfill
\subfigure{\includegraphics[width=0.085\columnwidth]{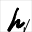}}
\hfill
\subfigure{\includegraphics[width=0.085\columnwidth]{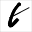}}
\hfill
\subfigure{\includegraphics[width=0.085\columnwidth]{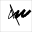}}
\hfill
\subfigure{\includegraphics[width=0.085\columnwidth]{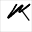}}
\hfill
\subfigure{\includegraphics[width=0.085\columnwidth]{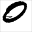}}
\hfill
\subfigure{\includegraphics[width=0.085\columnwidth]{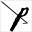}}
\hfill
\subfigure{\includegraphics[width=0.085\columnwidth]{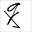}}
\hfill
\subfigure{\includegraphics[width=0.085\columnwidth]{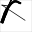}}
\hfill
\subfigure{\includegraphics[width=0.085\columnwidth]{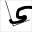}}
\hfill
\subfigure{\includegraphics[width=0.085\columnwidth]{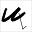}}
\hfill
\subfigure{\includegraphics[width=0.085\columnwidth]{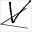}}
\hfill
\subfigure{\includegraphics[width=0.085\columnwidth]{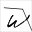}}
\hfill
\subfigure{\includegraphics[width=0.085\columnwidth]{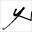}}
\hfill
\subfigure{\includegraphics[width=0.085\columnwidth]{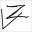}}
\caption{Sample input for each of the 20 single-stroke-characters in this task: a, b, c, d, e, g, h, l, m, n, o, p, q, r, s, u, v, w, y, z. Variations like the added random lines make this task challenging.}
\label{fig:characters}
\end{figure}

In this experiment, we test learning with side information for handwritten character recognition in images (see Fig.~\ref{fig:characters}), where we use the pen trajectory as side information. As in the previous experiment, we vary (i)~the representation of the side information, either as continuous vectors or discrete categories; (ii)~the pattern: direct, multi-task, or multi-view; and (iii)~the training procedure: decoupled, pre-train and finetune, or simultaneous. In all our experiments, we keep the number of unlabeled data and side information fixed and examine how the accuracy of the main task is affected by changing the number of labeled data. All approaches use the same convolutional neural network architecture. As baselines we use purely supervised learning and unsupervised pre-training (deep auto-encoder).

\begin{figure}[H]
\centering
\subfigure[Continuous side information\label{subfig:exp_handwritten_continuous}]{\includegraphics[width=0.8\columnwidth]{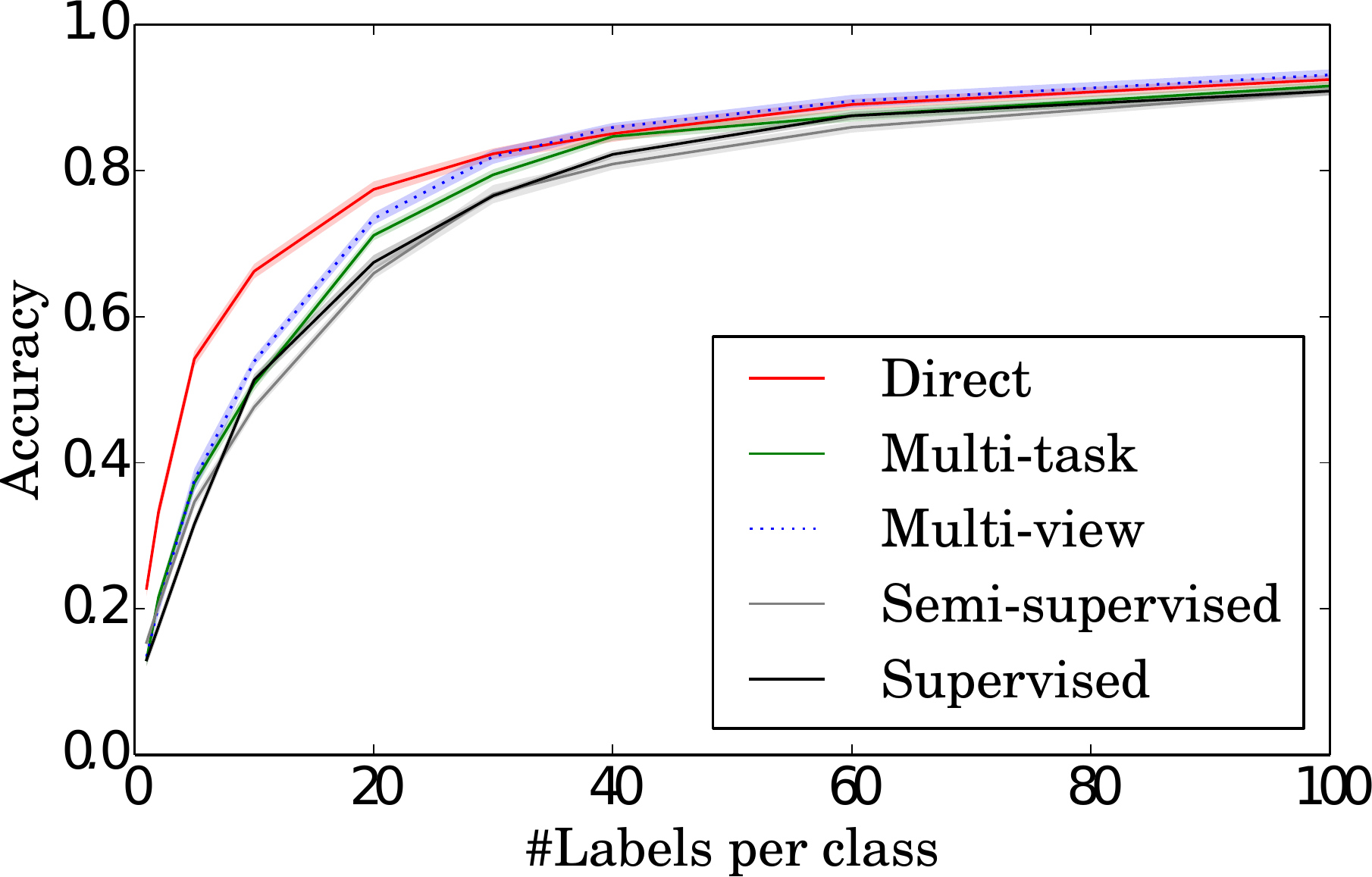}}
\\
\subfigure[Discretized side information\label{subfig:exp_handwritten_discrete}]{\includegraphics[width=0.8\columnwidth]{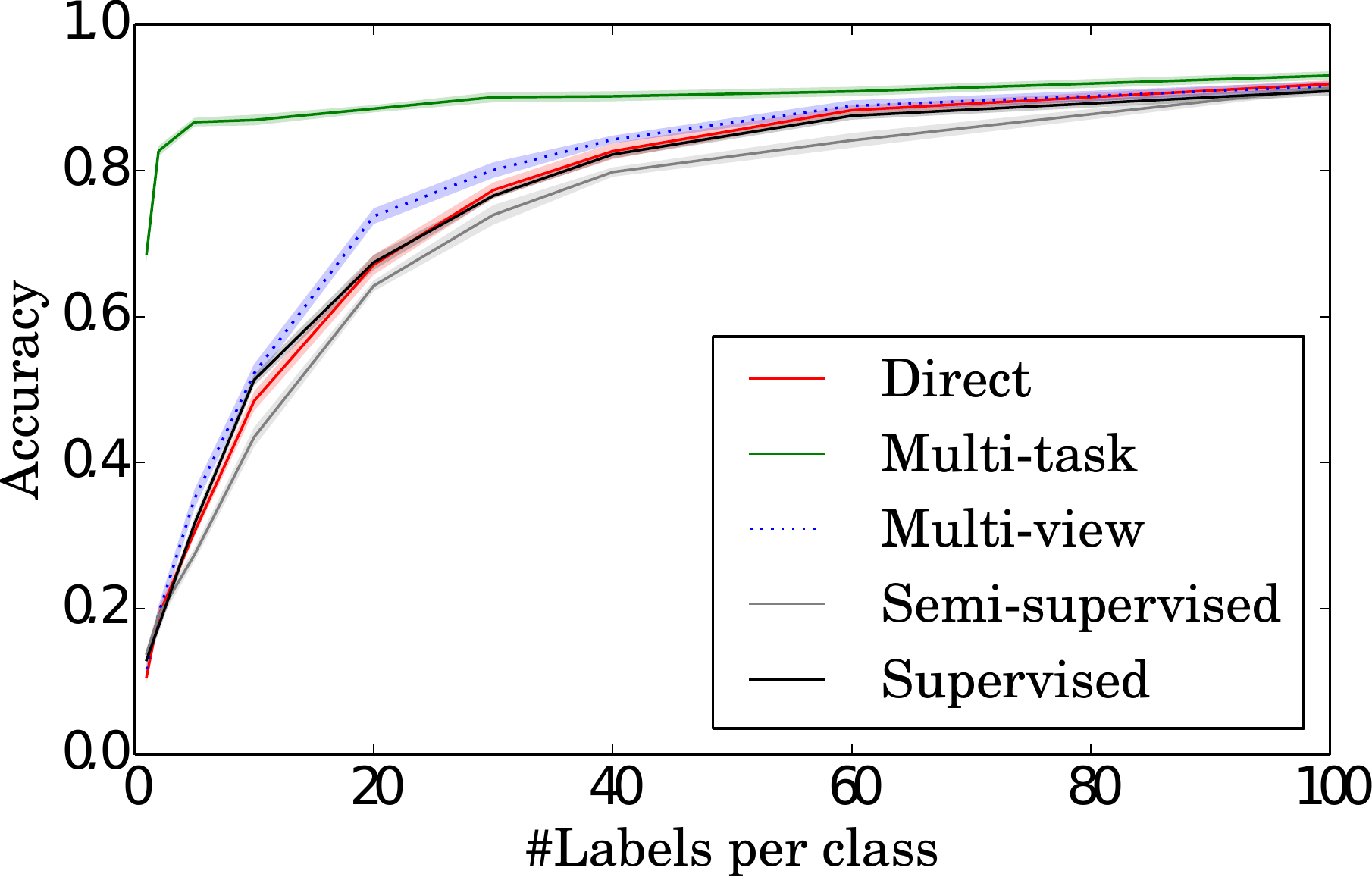}}
\\
\subfigure[Training procedures\label{subfig:exp_handwritten_training}]{\includegraphics[width=0.8\columnwidth]{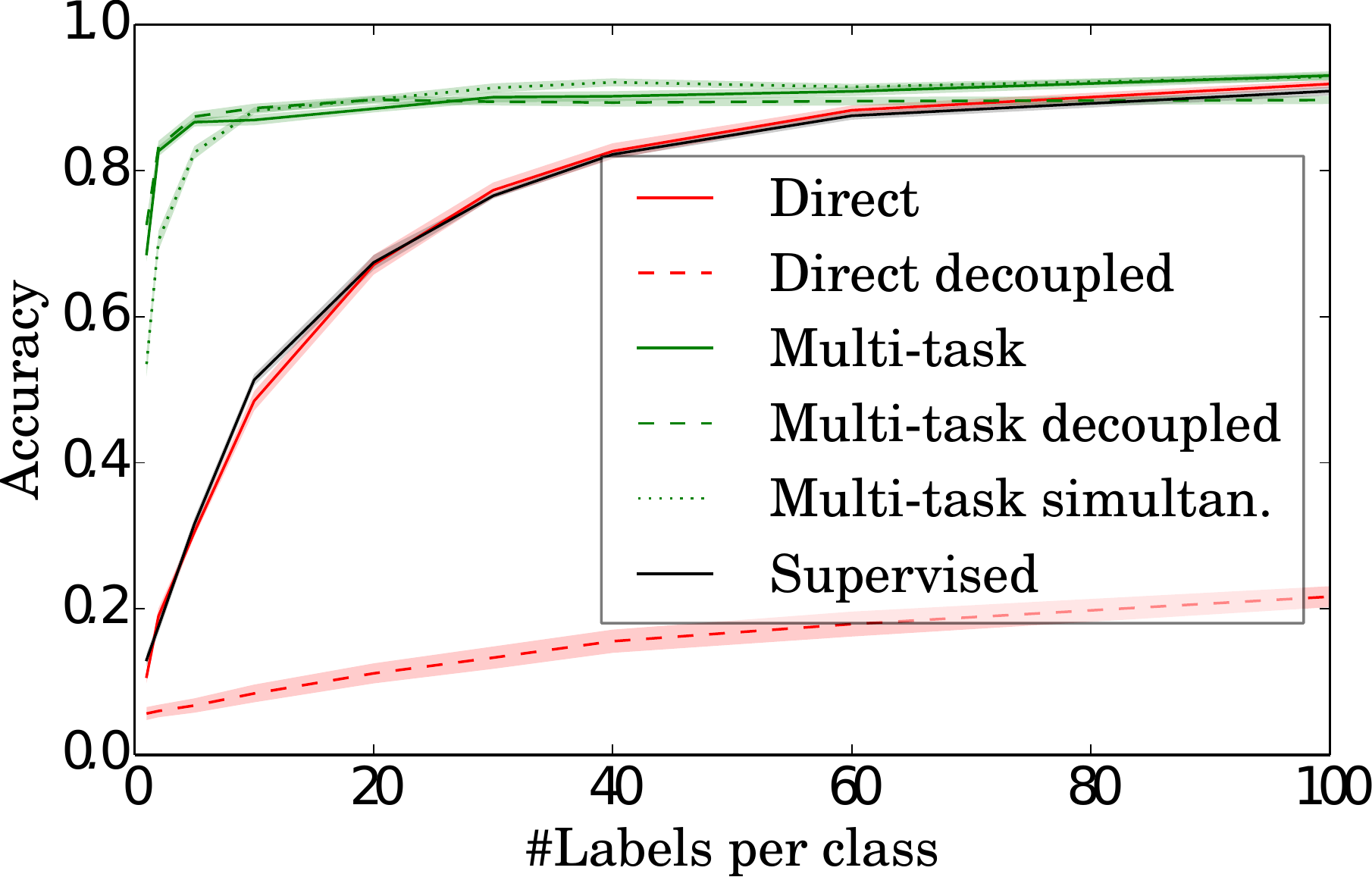}}
\caption{Results for handwritten character recognition task. (a) uses continuous side information. (b) and (c) use the discretized side information. Line styles denote the training procedure (solid = pretrain/finetune, dashed = decoupled training, dotted lines = simultaneous training).}
\label{fig:exp_handwritten}
\end{figure}

\tightparagraph{Results} First, we see that learning with side information can dramatically improve generalization, achieving the same performance using 5 labels per class as the baselines achieve with 100 (see Fig.~\ref{subfig:exp_handwritten_discrete}). Second, for this task, only the multi-task pattern exhibits this significant increase in performance. However, we also note that the effectiveness of learning with side information does not only depend on the pattern, but also on the representation of the side information (compare Figs.~\ref{subfig:exp_handwritten_continuous} and \ref{subfig:exp_handwritten_discrete}):
When we use the vector of pen coordinates as side information, the direct pattern provides some improvement over the baselines, but the multi-task and multi-view patterns do not improve the performance by large amounts (see Fig.~\ref{subfig:exp_handwritten_continuous}). However, when discretizing the trajectories into a small number of categories and applying the multi-task pattern for predicting these categories, we drastically reduce the number of labels required to solve this task (see Fig.~\ref{subfig:exp_handwritten_discrete}). This is not the case if we use other patterns. Moreover, we see that multi-task learning applied to the discretized side information is not significantly influenced by the training procedure (compare to direct pattern). This shows how---in this task---the multi-task pattern is able to find a good intermediate representation independently of the main objective (Fig.~\ref{subfig:exp_handwritten_training}). These results confirm our hypothesis that the available side information must match the prior imposed by the applied pattern in order to improve generalization.

\section{Conclusion and Discussion}

In this paper, we show how learning with side information provides a new perspective on machine learning, and complements existing paradigms such as supervised learning, representation learning, and deep learning. This new perspective allows us to connect various methods in the literature that (implicitly) use side information. It also enables us to systematically analyze these methods and extract patterns from them that show how they incorporate different priors about how side information relates to the target function. Since different priors coincide with different learning tasks, we hypothesize that the performance of these patterns will vary strongly depending on the applicability of the corresponding prior. Our experiments confirm this hypothesis and also show that learning with side information can substantially improve generalization.

Our perspective of learning with side information can be helpful in a number of ways. First of all, the patterns that we have presented allow researchers and practitioners to exploit side information in novel tasks in a systematic fashion. Applying the patterns in novel tasks is facilitated by our publicly available implementation and a broad overview of existing methods and applications in this paper. Our literature review provides a formalization that unifies different lines of research, which currently seem to be unaware of the strong similarities among them. This common view will allow researchers to exchange ideas more easily, to find novel patterns for using side information, and to effectively combine patterns to exploit multiple sources of side information.

Moreover, we expect learning with side information to be very effective beyond supervised learning settings. In particular, reinforcement learning can benefit significantly
because of the strong relationship between the different data sources (observations, actions, and rewards over time). By formulating priors over their relationships, we can exploit this rich side information in order to learn better state, action, and policy representations. This stands in contrast to most datasets available in machine learning, which are shaped according to the supervised learning paradigm and thus only consist of input/output samples. We, therefore, suggest to construct and augment datasets with relevant side information.

Although the utility of the view that we have presented can ultimately only be estimated in hindsight, we strongly believe that unifying ideas and providing new perspectives is vital to scientific progress, as exemplified by~\citet{bengio_representation_2013}. We hope that the presented perspective of learning with side information triggers further research in that direction that generates new insights in our field.

\subsubsection*{Acknowledgments}

We gratefully acknowledge the funding provided by the European Commission (SOMA project, \mbox{H2020-ICT-645599}), the German Research Foundation (Exploration Challenge, \mbox{BR 2248/3-1}), and the Alexander von Humboldt foundation 
(funded by the German Federal Ministry of Education and Research). We would like to thank Marc Toussaint and the University of Stuttgart for granting us access to their GPU cluster, and Sven D\"ahne, George Konidaris, Johannes Kulick, Tobias Lang, Robert Lieck, Ingmar Posner, and Michael Schneider for fruitful discussions and comments on this manuscript.

\bibliography{cml_paper}
\bibliographystyle{icml2016}

\clearpage

\begin{appendix}

\section{Patterns as Probabilistic Graphical Models}
\label{app: pgm}

To complement the computation flow schemas of the patterns used throughout the paper, we provide an interpretation of the main patterns as probabilistic graphical models (PGMs). These models treat the variables and functions introduced in Sec.~\ref{sec: cml} as random variables, represented as nodes. Arrows between these random variables indicate causal relationships. Gray nodes indicate observable, and white nodes latent random variables. The latent functions can be learned by performing inference in these models.

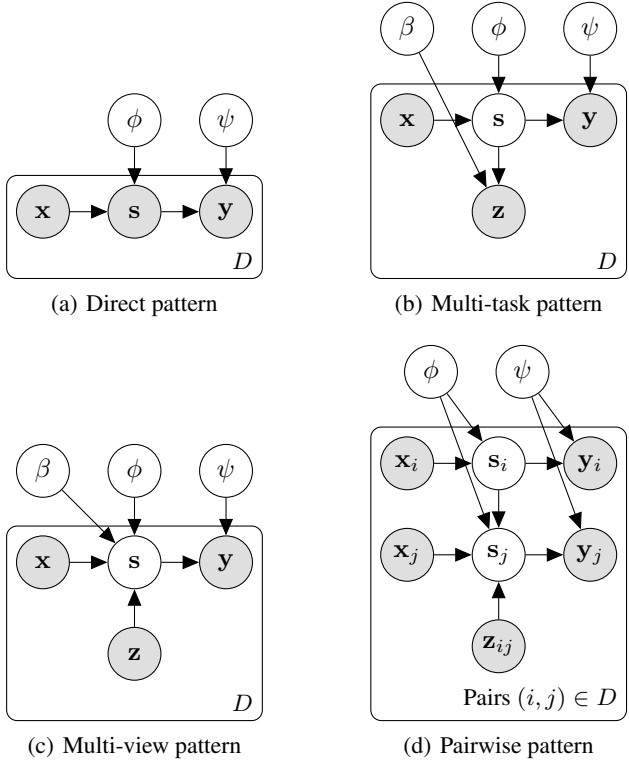
\begin{figure}[H]
\centering
\center
\subfigure[Direct pattern]{\begin{tikzpicture}
\label{fig:pgm_pattern_direct}

  \node[obs]                               (x) {$\varx$};
  \node[obs, right=0.5cm of x]               (s) {$\vars$};
  \node[latent, above=0.5cm of s] (phi) {$\funcphi$};  
  
  \node[obs, right=0.5cm of s]               (y) {$\vary$};
  \node[latent, above=0.5cm of y] (psi) {$\funcpsi$};
  
 
  \edge {x,phi} {s};
  \edge {s,psi} {y};

  \plate {xsyc} {(x)(y)(s)} {$D$};

\end{tikzpicture}}
\hfill
\subfigure[Multi-task pattern]{\begin{tikzpicture}
\label{fig:pgm_pattern_multitask}
  \node[obs]                               (x) {$\varx$};
  \node[latent, right=0.5cm of x]               (s) {$\vars$};
  \node[latent, above=0.5cm of s] (phi) {$\funcphi$};  
  
  \node[obs, right=0.5cm of s]               (y) {$\vary$};
  \node[latent, above=0.5cm of y] (psi) {$\funcpsi$};
  
  \node[obs, below=0.5cm of s] (c) {$\varc$};
  \node[latent, above=0.5cm of x] (beta) {$\funcchi$};
 
  \edge {x,phi} {s};
  \edge {s,psi} {y};
  \edge {s,beta} {c};

  \plate {xsyc} {(x)(y)(s)(c)} {$D$};

\end{tikzpicture}}
\\
\subfigure[Multi-view pattern]{\begin{tikzpicture}
\label{fig:pgm_pattern_multiview}
  \node[obs]                               (x) {$\varx$};
  \node[latent, right=0.5cm of x]               (s) {$\vars$};
  \node[latent, above=0.5cm of s] (phi) {$\funcphi$};  
  
  \node[obs, right=0.5cm of s]               (y) {$\vary$};
  \node[latent, above=0.5cm of y] (psi) {$\funcpsi$};
  
  \node[obs, below=0.5cm of s] (c) {$\varc$};
  \node[latent, above=0.5cm of x] (beta) {$\funcchi$};
 
  \edge {x,phi} {s};
  \edge {s,psi} {y};
  \edge {c,beta} {s};

  \plate {xsyc} {(x)(y)(s)(c)} {$D$};

\end{tikzpicture}}
\hfill
\subfigure[Pairwise pattern]{\begin{tikzpicture}
\label{fig:pgm_pattern_pairwise}
  \node[obs]                               (xi) {$\varx_i$};
  \node[latent, right=0.5cm of xi]               (si) {$\vars_i$};
  \node[obs, right=0.5cm of si]               (yi) {$\vary_i$};
  
  \node[obs, below=0.5cm of xi]                               (xj) {$\varx_j$};
  \node[latent, right=0.5cm of xj]               (sj) {$\vars_j$};
  \node[obs, right=0.5cm of sj]               (yj) {$\vary_j$};
  
  \node[obs, below=0.5cm of sj] (cij) {$\varc_{ij}$};
  
  \node[latent, above=0.5cm of si, xshift=-0.9cm] (phi) {$\funcphi$}; 
  \node[latent, above=0.5cm of yi, xshift=-0.9cm] (psi) {$\funcpsi$};
 
  \edge {xi,phi} {si};
  \edge {si,psi} {yi};
  \edge {xj,phi} {sj};
  \edge {sj,psi} {yj};
  \edge {si,cij} {sj};

  \plate {xsyc} {(xi)(yi)(cij)} {$\text{Pairs } (i,j) \in D$};
\end{tikzpicture}}
\caption{Probabilistic graphical models for patterns
\label{fig:pgm_pattern}
}
\end{figure}

The PGMs for the four main patterns are shown in Fig.~\ref{fig:pgm_pattern}. The variables $\varx, \vars$, $\varc$ and $\vary$ are observable random variables and are part of the training data $D$. The functions $\funcphi$, $\funcpsi$ and $\funcchi$ have become latent random variables, which the observed variables are conditioned on. We now discuss aspects of individual patterns.


The \emph{direct pattern} is shown in Fig.~\ref{fig:pgm_pattern_direct}. In comparison to its computation flow graph (Fig.~\ref{fig:pattern_direct}), the side information $\varc$ is considered as drawn from the distribution over $\vars$, and therefore $\varc$ does not appear in the graphical model of the direct pattern.

Fig.~\ref{fig:pgm_pattern_multitask} and Fig.~\ref{fig:pgm_pattern_multiview} show the PGM for the \emph{multi-task} and \emph{multi-view pattern}, respectively (computation flow graphs: Fig.~\ref{fig:pattern_multitask} and Fig.~\ref{fig:pattern_correlation}). We see that they are structurally similar and only differ on whether $\varc$ depends on $\vars$ and $\funcchi$ or whether $\vars$ depends on $\varc$ and $\funcchi$. In this regard, they are equivalent to their corresponding computation flow schemas apart from the fact that the PGM for the multi-view pattern conceals how the variables $\varx$ and $\varc$ belong to the functions $\funcphi$ and $\funcchi$.


Finally, the prototypical \emph{pairwise pattern} is shown in Fig.~\ref{fig:pgm_pattern_direct} (computation flow graph: Fig.~\ref{fig:pattern_pairwise}). Notice that here $\vars_j$ is conditioned on  $\varc_{i,j}$ and $\vars_i$, reflecting the fact that $\varc_{i,j}$ is information about how $\vars_j$ relates to $\vars_i$. 

Several interesting research questions arise from the probabilistic view on learning with side information. The majority of the reviewed literature uses non-probabilistic loss functions, mostly for training neural networks. Translating them into probabilistic ones is an interesting, but non-trivial research question, as the recent work on variational auto-encoders (which are a probabilistic version of auto-encoders) shows \citep{kingma_auto-encoding_2014}. A similar question arises on the relationship of the side objectives and prior probability distributions on $\funcphi, \funcpsi$, $\funcchi$ and $\vars$. It would be interesting to investigate whether certain side objectives can be shown to be equivalent to priors in the Bayesian sense, similar to the well-known fact that L2 regularization is equivalent to a Gaussian prior.





\section{Experimental Methods}
\label{app: experiments}

The implementation of our experiments is based on Theano~\citep{bastien_theano:_2012} and Lasagne\footnote{\url{https://github.com/Lasagne/Lasagne}}. We have made our code publicly available at: 
[url removed for double-blind review]

\subsection{Synthetic Task}
\label{app: synthetic task}

\tightparagraph{Task}
The task consists of an agent moving randomly through a 1-dimensional state space $s_t \in \mathbb{R}$ where $t$ denotes the time index.  The space is split into two regions $O_1 = \{ s\ |\ s > 0\}$ and $O_2 = \{ s\ |\ s < 0\}$ and the goal of the learner is to determine in which of the two regions the agent is located. However, the learner cannot observe the state space directly, it only gets $d$-dimensional observations which are obfuscated by $d-1$ distractors $u_t^{(i)}$, $i \in \{1, \ldots, d-1\}$: the observation is generated by  applying a random rotation $\mathbf{R}$ to the concatenated state and distractor vector: $\varx_t = \mathbf{R}\ [ s_t, u_{t}^{(1)}, \ldots, u_{t}^{(d-1)} ]$. 
In every time step, both the state as well as the distractor dimensions change randomly: 
$s_{t+1} = s_t + \varepsilon_t^{(s)}$, $u_{t+1}^{(i)} = u_{t}^{(i)} + \varepsilon_t^{(i)}$, where $\varepsilon_t^{(s)}, \varepsilon_t^{(i)} \sim \mathcal{N}(0,1)$. 
In addition the agent receives a supervised signal $\vary$ which is 0 if the agent is in region $O_1$ and 1 in zero $O_2$.

\tightparagraph{Baselines} 
We compare different variants of learning with side information to a supervised method (logistic regression mapping $\varx$ directly to $\vary$) and to two semi-supervised methods. For the semi-supervised baselines we apply either PCA or SFA to learn a 1-dimensional $\vars$, and then train a logistic regression mapping from $\vars$ to $\vary$. For the logistic regression, we use L$_2$ regularization with $C \in \{ 0.001, 0.01,$ $0.1, 1.0, \infty \}$ and choose the result with the lowest test error. (We do not apply L$_2$ regularization for variants of learning with side information.)

\tightparagraph{Patterns} 
We implemented the four principal patterns from Section~\ref{sec:patterns}, using a linear function for $\funcphi(\varx) = \mathbf{w}_\funcphi^T \varx = \vars$ and a logistic function for $\funcpsi(\vars) = \frac{1}{1+e^{-\mathbf{w}_\funcpsi^T \vars}}$. We apply Stochastic Gradient Descent with Nesterov momentum with value $0.9$ to learn $\funcphi$ and $\funcpsi$ using a logistic regression loss in addition to the side objective.
We train each pattern using the decoupled and simultaneous training procedures (pre-train/fine-tune is futile due since $\funcphi$ and $\funcpsi$ are linear, and both the target and the side objectives are convex; see Section~\ref{subsec: training procedures}).\\
For the \textbf{direct pattern}, we learn $\funcphi$ directly by performing a linear regression on $\varc$, using the mean squared error loss. When training simultaneously, we weigh the main and the side objective equally.\\
We implement the \textbf{multi-task pattern} by using a linear function $\funcchi(\vars) = \mathbf{w}_\funcchi^T \vars$ for the auxiliary task and optimize it using linear regression. Again, we weigh the main and the side objective equally.\\
We implement two versions of the \textbf{multi-view pattern}: first, the \textbf{correlation} variant, using $\funcchi(\varc) = \mathbf{w}_\funcchi^T \varc = \vars'$, optimizing for $\vars \approx \vars'$, secondly, the \textbf{prediction} variant. For the correlation pattern we have to give weight $0.99$ to the supervised and $0.01$ to the side objective, since the gradients from the side objective (MSE loss) and the supervised softmax loss differ by several orders of magnitude.\\
Finally, we implement the \textbf{pairwise pattern}, in form of the transformation pattern. We use a simplified version of the variant depicted in Fig.~\ref{fig:transf_pred_transf} with a fixed auxiliary function $\funcchi(\vars_i, \vars_{j}) = \vars_i - \vars_{j}$. Again, we weigh the main and the side objective equally.

%
%

We evaluate subsets of the implemented patterns with three types of side information. In each experiment, we use different amounts of $(\varx, \vary, \varc)$ triplets for learning, and test the prediction accuracy in the main task for a test set of size 50000. The dimensionality of $\varx$ is set to 50. We average the results over 10 independently generated training and test sets.

\tightparagraph{Direct Side Information} 
First, we provide the learner with very informative data in the form of a noisy variant of the real state: $\varc_{t}^{(s)} = s_t + \varepsilon_t^{(s)}$, with $\varepsilon_t^{(s)} \sim \mathcal{N}(0,0.05)$. Figure~\ref{fig: toy direct} shows that all variants of learning with side information except for the multi-view-prediction pattern generalize well, even given low amounts of data. The unsupervised methods fail to extract a good state representation since the real hidden $\vars$ neither exhibits high variance, nor a slower trajectory than the distractors. Pure supervised logistic regression works better, but even when doubling the number of $(\varx, \vary)$ pairs, it does not reach the performance of the methods that use side information.
We believe that the multi-view-prediction pattern works badly because it does not propagate enough information from the learned $\funcpsi$ to regularize $\funcphi$.

\tightparagraph{Embedded Side Information} 
In the second experiment, we provide side information corresponding to an additional, noisy sensor view by mapping the state into a different $e$-dimensional observation space, $\varc_{t}^{(v)} = \mathbf{Q}\ [ s_t + \varepsilon_t^{(v)}, v_{t}^{(1)},$ $\ldots, v_{t}^{(e-1)} ]$ with distractors $v_t^{(i)}$,  random rotation matrix $\mathbf{Q}$ 
and $\varepsilon_t^{(v)} \sim \mathcal{N}(0,0.05)$. In the experiments, we set $e = \frac{d}{2}$.
Figure~\ref{fig: toy embedded} shows that most variants of learning with side information still outperform the supervised method, but need more data to generalize well due to the less informative side information. The multi-view method performs best, whereas the multi-task performs even worse than logistic regression. Moreover, we see that the simultaneous training procedures outperform the decoupled variants, most drastically in the multi-view pattern.

\tightparagraph{Relative Side Information} 
The last type of data corresponds to a noisy variant of the ``actions'', $\varc_{t}^{(a)} = s_t - s_{t-1} + \varepsilon_t^{(a)}$ with $\varepsilon_t^{(a)} \sim \mathcal{N}(0,0.05)$. Fig.~\ref{fig: toy relative} shows clearly that this side information is highly useful, and allows to learn from few samples.

\subsection{Handwritten Character Recognition}

\tightparagraph{Dataset and task} The dataset for this experiment is based on the character trajectories dataset\footnote{\url{https://archive.ics.uci.edu/ml/datasets/Character+Trajectories}}, which consists of time series of pen velocities in x and y direction and pen tip force (more details in \citet{williams_primitive_2007}). The dataset includes 20 characters that can be written in a single stroke. Based on this dataset, we generate monochrome images of size $32\times32$ pixels. During image generation, we add different variations to make this task more challenging. We trace the character trajectories with varying pen width, we translate the characters randomly, and we overlay distracting lines that connect three random points in the image (see Fig.~\ref{fig:characters}). These images form the input $\varx$ for this task. Additionally, we generate side information in the form of coordinates of 32 points along the character trajectory making up a 64D vector $\varc$. Unlike the input images, these points are not translated. The task is to recognize which of the 20 characters is in the given image. The training data consist of 100 input/side information pairs per character, a random subset of which are labeled. In our experiment, we vary the number of labels per character from 1 to 100.

\tightparagraph{Neural network structure and training} We use a convolutional neural network (CNN) with rectified linear units (ReLU). We first apply a convolution with 32 filters of size $5\times5$ followed by ReLU non-linearity and max-pooling. The same sequence is repeated, followed by $50\%$ dropout and a fully connected layer of 32 ReLUs (the intermediate representation $\vars$). This is again followed by a $50\%$ dropout and 20 softmax output units. The entire network has 52756 parameters. The supervised task is formulated using the categorical crossentropy loss and optimized using Nesterov momentum with learning rate 0.003, momentum 0.9, and batch size 20 for 100 epochs, followed by 10 epochs with learning rate 0.0003. All experiments are repeated 10 times.

\tightparagraph{Discretization} In the experiment, we test two different representations of the side information. The original continuous vector representation and a discretization into 32 classes, which we obtain with k-means clustering. The rationale behind the discretization is that the exact trajectory cannot be recovered from the image (because it is not clear from the image where the character trajectory starts). We tested the same discretization on the image in the semi-supervised baseline.

\tightparagraph{Applied patterns} We compare the direct pattern, the multi-task pattern, and the multi-view pattern. The direct pattern uses a mean-squared-error objective to enforce the intermediate representation to be equal to a 32D version of the pen trajectory or the one-hot-vector that corresponds to the discretized trajectory. The multi-task pattern incorporates an additional network layer to predict the side information, either a linear layer with mean-squared-error loss to predict the trajectory or a softmax layer with cross-entropy loss to predict its discretization. The multi-view pattern uses two ReLU-layers with 32 units and dropout to compute the intermediate representation $\vars'$ which we tie to $\vars$ with a mean-squared-error objective. Since this objective creates trivial solutions if trained independently, we optimize it simultaneously with the main objective (weighing them with $0.05$ and $0.95$, respectively). All other patterns are trained using the pretrain/finetune procedure unless indicated otherwise. For simultaneous training of the multi-task pattern, we used uniform weighting.

\tightparagraph{Baselines} We compare against supervised learning on the labeled data and semi-supervised baselines: In the continuous case, we reconstruct the image using a convolutional autoencoder that mirrors the structure of the convolutional network. In the discrete case we use a similar structure as for multi-task learning but predict the discretized image instead of the discretized trajectory.

\section{Overview of Related Work}

\label{app: table}

In the following table, we summarize related works that apply learning with side information. Since an exhaustive list of references for each pattern is beyond the scope of this paper, we include works that span a wide variety of instantiations of the proposed patterns and refer to survey articles if available.

{ \footnotesize
Abbreviations: AE=auto-encoder, CCA=canonical correlation analysis, ED=eigen decomposition, GMLVQ=generalized matrix learning vector quantization, kNN=k-nearest-neighbors,
LBP=locally binary pattern, MMD=maximum mean discrepancy, NN=neural network, RBM=restricted Boltzmann machine, RL=reinforcement learning, SGD=stochastic gradient descent, SL=supervised learning (classification unless stated otherwise), SVM=support vector machine, UL=unsupervised learning
}

\onecolumn



%

\newcolumntype{L}[1]{>{\raggedright\let\newline\\\arraybackslash\hspace{0pt}}m{#1}}
\newcolumntype{C}[1]{>{\centering\let\newline\\\arraybackslash\hspace{0pt}}m{#1}}
\newcolumntype{R}[1]{>{\raggedleft\let\newline\\\arraybackslash\hspace{0pt}}m{#1}}

\newcommand{\decoup}{\emph{(decoupl.)}}
\newcommand{\simul}{\emph{(simul.)}}
\newcommand{\prefine}{\emph{(pre-train/fine-tune)}}


{
\scriptsize
\begin{longtable}{| C{.08\textwidth} | L{.16\textwidth} | L{.1\textwidth} | L{.13\textwidth} | L{.215\textwidth} | L{.12\textwidth} |} 
\hline
\textbf{Pattern}  
& \textbf{Side Objective} 
& \textbf{Articles} 
& \textbf{Method},\newline \textbf{Train. Procedure} 
& \textbf{Application: \newline Task, Input, Dataset} 
& \textbf{Side Information} 
 \\ 
\hline \hline 
\emph{Direct} \newline
(Fig.~\ref{fig:pattern_direct})
& SVM loss
& \citet{cheng_improved_2007} 
& SVM \decoup
& SL: Contact prediction on sequences
& Secondary (3D) structure categories
\\
\cline{2-6}
& Regression on highly predictive features of side information
& \citet{chen_boosting_2012} 
& AdaBoost+ \simul
& SL on images: Digit \citep{vapnik_new_2009}, facial expression (Cohn-Kanade)
& Holistic image descriptions, LBP features from high-res images
\\
\cline{2-6}
& Regression loss 
& \citet{vapnik_learning_2015}
& SVM with knowledge transfer \decoup
& SL on images 
\newline Theoretical analysis: learning using privileged information
& Image sections
\\ 
\hline \hline
\emph{Multi-task}\newline
(Fig.~\ref{fig:pattern_multitask})
& Various supervised: hinge, MSE, softmax
& \citet{caruana_multitask_1997}
& NN \simul
& SL: pneumonia detection
& Hematocrit, white blood cell count, potassium
\\ 
\cline{3-6}
&  
& \citet{evgeniou_regularized_2004}
& SVM \simul
& SL: exam score prediction
& One task per school
\\ 
\cline{3-6}
&  
& \citet{levine_end-end_2015}
& Conv. NN \decoup
& RL on RGB-D: robot manipulation
& Image class, object pose
\\ 
\cline{3-6}
&  
& \citet{zhao_improved_2015}
& Conv. NN \simul
& SL on images (YYY-20M) 
& Object pose
\\ 
\cline{3-6}
&  
& \citet{pan_survey_2010}
& Survey
& SL, UL
& -
\\ 
\cline{3-6}
& 
& \multicolumn{2}{l|}{
\citet{baxter_model_2000,ando_framework_2005}
}
& Theoretical analysis of
& -
\\ 
& 
& \multicolumn{2}{l|}{
\citet{maurer_bounds_2006}
}
& multi-task learning
& 
\\ 
\hline \hline
\emph{Multi-view}\newline
(Fig.~\ref{fig:pattern_correlation})
& Kernel CCA+soft margin SVM hinge loss
& \citet{farquhar_two_2005}
& SVM-2K \simul
& SL on images (PASCAL-VOC)
& Keypoint features (SIFT)
\\ 
\cline{2-6}
& AE reconstruction error
& \citet{ngiam_multimodal_2011}
& RBM / NN \simul
& SL on video/audio (various, e.g. CUAVE, AVLetters)
& Video/audio
\\ 
\cline{2-6}
& Adjusted rand index, mutual information
& \citet{feyereisl_privileged_2012}
& k-means
& UL on images: MNIST
& Poetic descriptions
\\ 
\cline{2-6}
& Kernel CCA [+MMD for domain adaption]
& \citet{chen_recognizing_2014}
& Kernel SVM on kernel descriptor features
& SL on RGB: gender (EURECOM, LFW-a), object (RGB-D O.D., Catech-256)
& RGB-D
\\ 
\cline{2-6}
& SPoC (non-linear CCA)
& \citet{dahne_finding_2014}
& Linear/quadratic, ED \decoup
& SL: Mental state prediction
& EEG diff. subject
\\ 
\cline{2-6}
& CCA+AE reconstruction error
& \citet{wang_deep_2015}
& Deep Canonically Correlated AE and others \simul
& SL on images (MNIST), speech (XRMB), word embedding (WMT2011)
& Noisy images, articulations, 2nd language
\\ 
\cline{2-6}
&  -
& \citet{sun_survey_2013}
& Survey
& SL, UL
& -
\\ 
\cline{2-6}
& -
& \multicolumn{2}{l|}{\citet{blum_combining_1998}}
& Theoretical analysis of
& -
\\ 
& 	
& \multicolumn{2}{l|}{\citet{wang_new_2010}}
& multi-view learning
& 
\\ 
\hline \hline
\emph{Pairwise Similarity}\newline
(Fig.~\ref{fig:pattern_pairwise})
& Slowness (first equation in Sec.~\ref{sec:similarity} with $\varc_{ij} = \indicator{j=i+1}$ and covariance constraints).
& \citet{wiskott_slow_2002, legenstein_reinforcement_2010}
& Linear/quadratic, ED \decoup
& RL on images: Navigation (physical simulation)
& Time index
\\ 
\cline{2-5} 
&  Equations in Sec.~\ref{sec:similarity} with margin-based $\sigma(d)$ (see Section~\ref{sec:similarity})
& \citet{hadsell_dimensionality_2006}
& Conv. NN, \decoup
& UL on images: dimensionality reduction (MNIST, NORB)
& 
\\
\cline{3-5} 
& 
& \citet{weston_deep_2012}
& Conv. NN \simul
& SL on images (MNIST, COIL100)
& 
\\ 
\cline{2-5} 
& State predictability +variational AE
& \citet{watter_embed_2015}
& CNN \simul
& RL: inverted pendulum, cart-pole, robot arm
& 
\\ 
\cline{2-6}
& Adapted SVM loss 
& \citet{vapnik_new_2009}
& SVM with similarity control \simul
& SL: protein classification, finance market prediction, digit recognition
& 3D protein structure, future events, textual description
\\ 
\cline{2-6}
& Distance metric learning
& \citet{fouad_incorporating_2013}
& GMLVQ/kNN \decoup
& SL: images (MNIST); galaxy morphology
& Poetic descriptions; spectral features
\\ 
\cline{2-6}
(Fig.~\ref{fig:pattern_pairwise_label})
& Hierarchical multi-class loss
& \citet{silla_jr_survey_2010}
& \multicolumn{2}{l|}{Survery on hierarchical classification (SL)}
& Label similarity
\\ 
\hline \hline
\emph{Pairwise Transformation} 
(Fig.~\ref{fig:transf_pred_input})
& Softmax (?)
& \citet{hinton_transforming_2011}
& NN; transforming AE \simul
& SL for pose prediction (MNIST, 3D simulation)
& Relative pose
\\ 
\cline{2-6}
(Fig.~\ref{fig:transf_pred_state})
& See~\citep{hadsell_dimensionality_2006}
& \citet{jayaraman_learning_2015}
& Siamese-style conv. NN \simul
& SL on images (NORB, KITTI, SUN)
& Relative pose (discretized; with k-means)
\\ 
\cline{2-6}
(Fig.~\ref{fig:transf_pred_transf})
& Softmax
& \citet{agrawal_learning_2015}
& Siamese-style conv. NN \prefine
& SL on images (MNIST, SF, KITTI)
& Relative pose (discretized; uniformly)
\\ 
\cline{2-6}
& Various
& \citet{jonschkowski_learning_2015}
& Linear, SGD \decoup
& RL: control, navigation, 
& Actions, rewards, time
\\ 
\hline \hline
\emph{Irrelevance}\newline
(Fig.~\ref{fig:irrelevant})
& $\symloss \approx ||\funcpsi^T \funcchi||^2_\textrm{F}$, 
with $\funcpsi$, $\funcchi$ linear, $||\cdot||^2_\textrm{F}$ Frobenius norm.
& \citet{romera-paredes_exploiting_2012}
& Linear, orthogonal matrix factorization
& SL on images: emotion detection (JAFFE)
& Subject identity 
\\ 
\hline
\end{longtable} 
}

\end{appendix}

\end{document}